\documentclass[sn-mathphys]{sn-jnl}

\usepackage{multirow}
\usepackage{tabularx}
\usepackage{xcolor}
\usepackage[table]{xcolor}
\usepackage{colortbl}
\usepackage{tablefootnote}

\usepackage{caption}
\usepackage{cellspace}
\usepackage{makecell}
\usepackage{hyperref}
\usepackage{amsmath}
\usepackage{graphicx}
\usepackage{bbding}
\usepackage{url}
\usepackage{booktabs}
\usepackage{CJKutf8}
\usepackage{longtable}
\usepackage{array}
\usepackage{url}
\usepackage[numbers]{natbib}
\usepackage[dvipsnames]{xcolor}

\usepackage{threeparttable}

\usepackage[section]{placeins}

\jyear{2021}%

\theoremstyle{thmstyleone}%
\theoremstyle{thmstyletwo}%

\theoremstyle{thmstylethree}%

\begin{document}

\title[Survey on LLMs for Medical Reasoning]{\includegraphics[height=1cm]{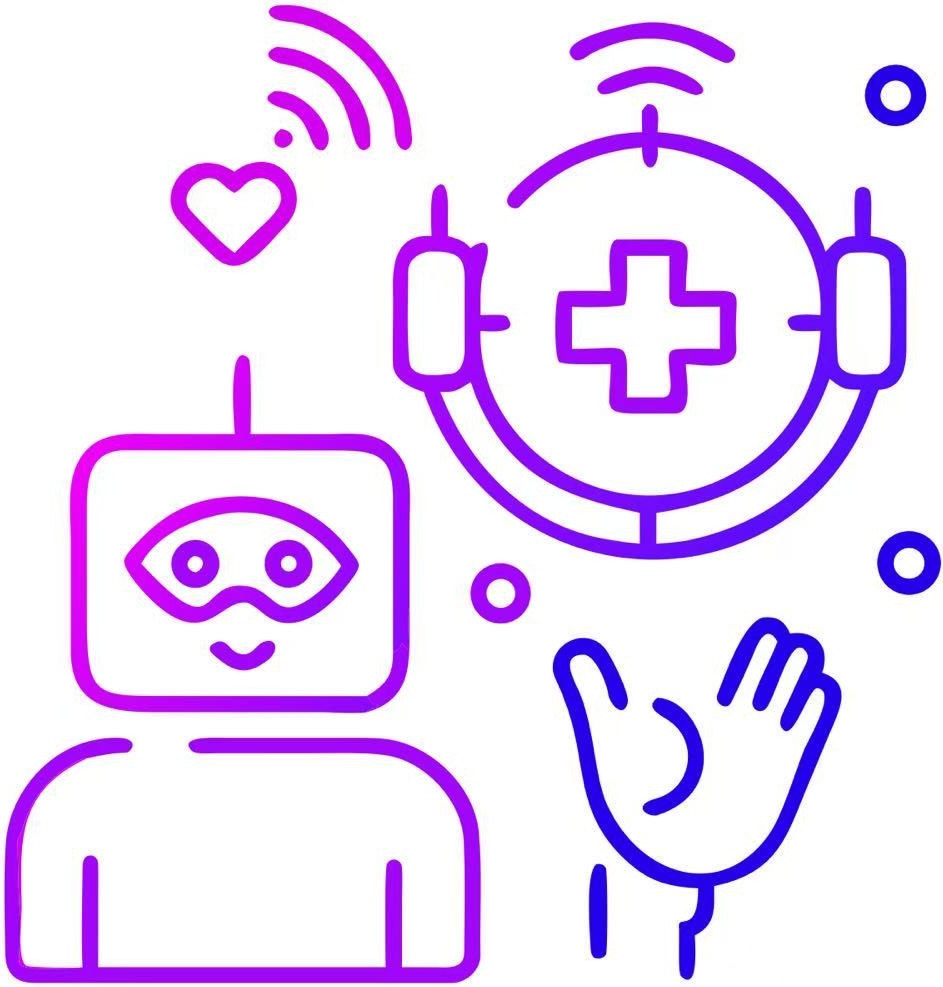} Aligning Clinical Needs and AI Capabilities: A Survey on LLMs for Medical Reasoning}

\insert\footins{\noindent\footnotesize $^1$ORCID of Qi Peng: https://orcid.org/0000-0001-7747-3293}
\insert\footins{\noindent\footnotesize $^*$ORCID of Changmeng Zheng: https://orcid.org/0000-0002-2945-8248}


\author[1,3]{Qi Peng}
\email{sepengqi@mail.scut.edu.cn}
\equalcont{These authors contributed equally to this work.}

\author[1]{Jiatong~Li}
\email{jiatong.li@connect.polyu.hk}
\equalcont{These authors contributed equally to this work.}

\author[1]{Sirui Huang}
\email{sirui.huang@connect.polyu.hk}
\equalcont{These authors contributed equally to this work.}

\author[1]{Yiyang Jiang}
\email{fletcher.jiang@connect.polyu.hk}

\author[2]{Kaisong~Gong}
\email{gks70853@connect.hku.hk}

\author[3]{Ronger~Ding}
\email{se30480335ding@mail.scut.edu.cn}

\author[4]{Shijie~Ye}
\email{yeshijie001119@gmail.com}

\author*[1]{Changmeng Zheng}
\email{changmeng.zheng@polyu.edu.hk}

\author[3]{Yi Cai}
\email{ycai@scut.edu.cn}

\author[5]{Xiaobo Yang}
\email{y110403606@126.com}

\author[6]{Jin Huang}
\email{michael\_huangjin@163.com}

\author[1]{Xiao-Yong Wei}
\email{cs007.wei@polyu.edu.hk}

\author[1]{Qing Li}
\email{qing-prof.li@polyu.edu.hk}

\affil[1]{\orgname{The Hong Kong Polytechnic University}, \orgaddress{\city{Hong Kong}, \country{China}}}

\affil[2]{\orgname{Hong Kong University}, \orgaddress{\city{Hong Kong}, \country{China}}}

\affil[3]{\orgname{South China University of  Technology}, \orgaddress{\city{Guangzhou}, \country{China}}}

\affil[4]{\orgname{University of Toronto}, \orgaddress{\city{Toronto}, \country{Canada}}}

\affil[5]{\orgname{Peking Union Medical College Hospital}, \orgaddress{\city{Beijing}, \country{China}}}

\affil[6]{\orgname{West China Hospital, Sichuan University}, \orgaddress{\city{Sichuan}, \country{China}}}




\abstract{Large language models (LLMs) have emerged as important tools in healthcare, showing growing potential for clinical reasoning and patient care. This survey examines recent progress in medical LLMs, focusing on reasoning applications and requirements. We present a dual-view approach that connects clinical practice with computational methods. On the clinical side, we establish a five-level competency scheme following Miller's Pyramid, progressing from knowledge recall to dynamic case management. On the computational side, we link deductive, inductive, and abductive reasoning patterns to common medical goals and tasks. We also introduce a benchmark dataset spanning five levels of medical reasoning capability and report results on 18 state-of-the-art models, revealing that medical specialist models excel in diagnosis-centric tasks while general models lead in decision support and dialogue. We conclude by discussing current progress and open challenges, including data limitations, hallucination, and grounding issues, and outline directions toward safer, more reliable, and workflow-ready systems.}




\maketitle

\begin{figure*}[!ht]
  \centering
  \includegraphics[width=\linewidth, keepaspectratio]{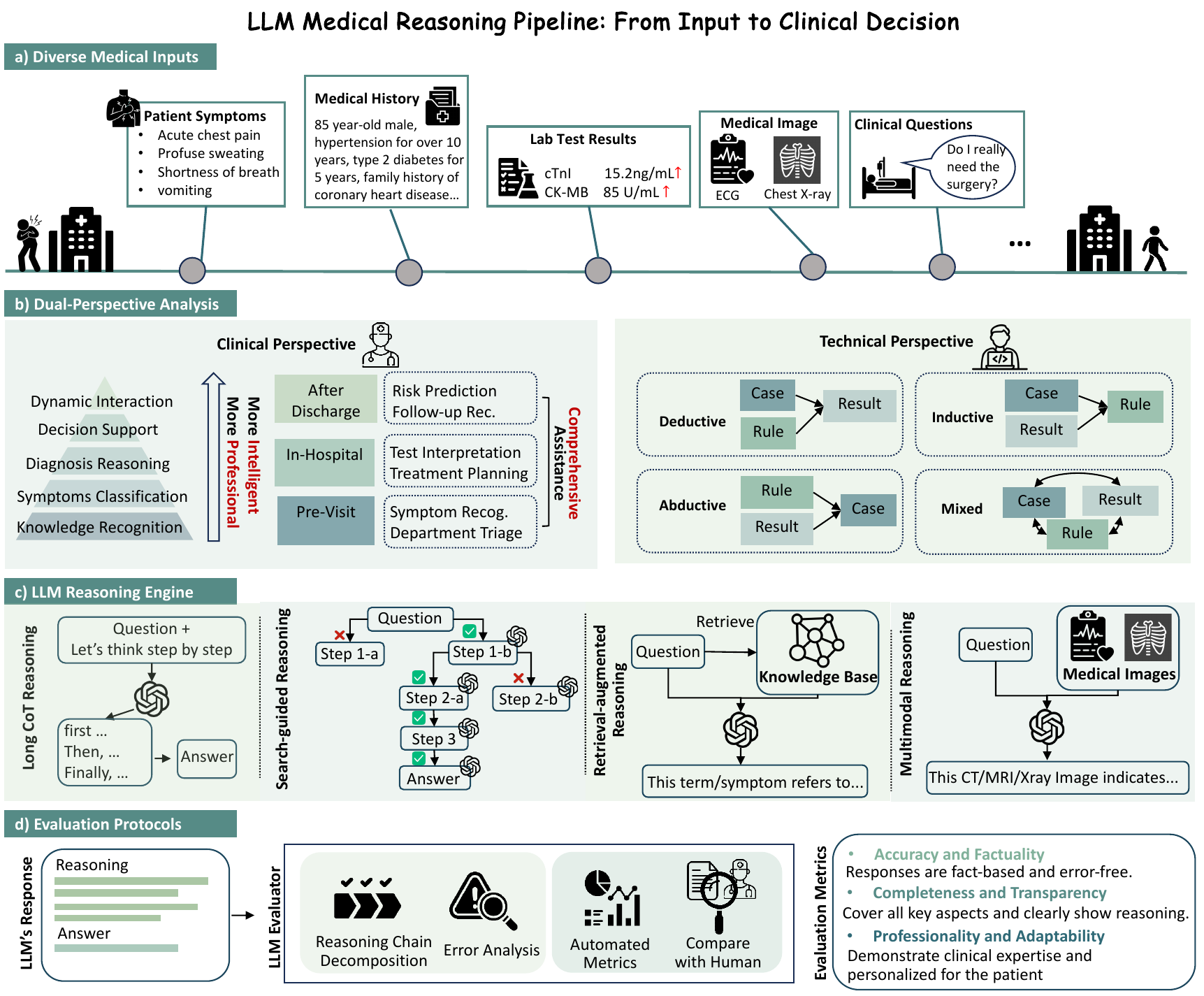}
  \caption{A general pipeline for LLM reasoning on medicine.}
  \label{fig:framework}
\end{figure*}

\section{Introduction}

The medical field faces growing pressure to deliver timely and accurate decisions while managing increasingly large volumes of data and constantly evolving evidence. Clinicians must skillfully combine patient history, physical examination findings, test results, and imaging data, while carefully weighing risks and preferences and acting under significant uncertainty. Although traditional tools \cite{aljabri2022towards,neves2014survey,kumar2013content,lin2025cohort} provide valuable support for data retrieval and rule execution, they often struggle when information is incomplete or conflicting, or when decisions depend critically on sequence, context, and complex trade-offs. Meanwhile, parallel advances in data availability and computational power have driven rapid progress in large language models \cite{A_survey_of_large_language_models,qiu2023pre}. These LLMs can effectively read and generate clinical text, connect to external sources, and engage in conversational interactions that naturally fit clinical workflows \cite{liang2025answer}. Consequently, this has led to widespread interest in applying LLMs across the entire care pathway, spanning from pre-visit triage and symptom assessment through diagnosis \cite{sun2024medical,shao2025mcanet,nie2025reparameterized}, treatment planning \cite{baig2024accuracy,sehar2026automatic}, documentation \cite{seo2024evaluation}, and follow-up care \cite{chow2016inter}.

This survey specifically focuses on medical reasoning capabilities rather than general language processing applications. Our primary goal is to \textit{clarify what kinds of reasoning are actually needed in clinical practice, how current model designs map to these specific needs, and how meaningful progress should be measured}. To achieve this, we adopt two complementary perspectives. \textbf{From the clinical view}, we systematically extend Miller's Pyramid \cite{The_assessment_of_clinical_skills/competence/performance} into five distinct levels: knowledge recognition and normalization; information classification and triage; causal reasoning and comprehensive diagnosis; decision support and individualized recommendations; and dynamic interaction in complex clinical scenarios. This framework provides a structured way to align specific tasks with expected competency levels along the care continuum. \textbf{From the computational view}, we organize existing methods by reasoning type and systematically link deductive, inductive, and abductive reasoning patterns to concrete medical objectives such as symptom normalization \cite{MedNorm}, risk prediction \cite{yang2024ensuring}, differential diagnosis \cite{hirosawa2023diagnostic}, and treatment selection \cite{baig2024accuracy}. Importantly, mixed reasoning settings that combine these patterns are common in realistic clinical workflows and are therefore treated explicitly in our analysis.

Building on these perspectives, we map datasets and benchmarks according to both reasoning type and clinical competency level, covering resources for entity grounding, clinical triage, diagnostic reasoning, question answering, imaging-informed tasks, documentation, and dialogue management. Furthermore, we thoroughly review model designs and training strategies currently used in practice, including instruction tuning with expert data \cite{zhang2023pmc}, chain-of-thought and long-form reasoning \cite{chen2025towards}, search- and tree-guided reasoning \cite{zhang2025llama}, retrieval-augmented generation with knowledge bases and literature \cite{bora2024systematic}, tool use and agent workflows \cite{liu2025autoct,zheng2024picture}, and multimodal integration for text and images \cite{boehm2022harnessing,feng2023towards}. We also discuss evaluation methodologies that extend beyond simple accuracy, encompassing factuality, reasoning completeness, internal consistency, evidence use, and uncertainty handling, utilizing both expert review and automated metrics.

To support standardized measurement across the field, we construct a comprehensive benchmark dataset with five levels of medical reasoning capability, containing 5{,}000 carefully curated samples (1{,}000 per level). Through systematic evaluation of 18 state-of-the-art models on this benchmark, we observe several important trends. Notably, medical specialist models tend to perform better on diagnosis-centric tasks, while high-capacity general models often excel in decision support, multi-turn dialogue, and summarization tasks. Interestingly, length-of-stay prediction and other structured temporal outcomes remain challenging for all systems tested. Our analysis reveals that model size alone does not fully explain performance differences; rather, instruction quality, domain-specific data, and reasoning training appear equally important. Based on these findings, a practical deployment approach would involve routing diagnosis-heavy queries to specialized medical models while leveraging general models for dialogue and supportive tasks.

This paper's contributions can be summarized as follows:
\begin{itemize}
\item We propose a dual-view framework that systematically links clinical competence levels with core reasoning types and aligns them with common medical tasks.
\item We introduce a standardized five-level benchmark with 5{,}000 samples and report detailed results on 18 state-of-the-art models, yielding insights on where medical and general models excel and where they struggle.
\item We provide a thorough review of training and inference methods for medical reasoning, including instruction tuning, chain-of-thought, retrieval and tool use, agent workflows, and multimodal integration, while outlining evaluation criteria that assess both answers and reasoning steps.
\item We systematically discuss current limitations and risks, including data gaps, hallucination, grounding challenges, and calibration issues, and suggest concrete directions for safer and more workflow-ready systems.
\end{itemize}

\textbf{Organization}: The remainder of this paper is structured as follows. Section~II provides background and presents our proposed taxonomy linking clinical competencies with reasoning types. Section~III systematically surveys datasets and benchmarks organized by reasoning type and clinical level. Section~IV reviews model designs and reasoning paradigms, including chain-of-thought, long-form, search-guided, retrieval-augmented, multimodal, and agentic methods. Section~V introduces our five-level benchmark and reports detailed results for 18 models with analysis of observed trends. Section~VI discusses current challenges and limitations. Section~VII outlines recent advances and future directions. Finally, Section~VIII concludes with key insights and implications.

\section{Background and Taxonomy}

The foundation of LLM reasoning in medicine rests upon three interconnected pillars: the evolution of large language models and their core technologies, the theoretical understanding of reasoning processes, and the established frameworks for assessing clinical competence. This section provides comprehensive background knowledge essential for understanding the current state and future potential of medical reasoning systems.

\subsection{Overview of Medical Large Language Models}

The landscape of artificial intelligence has been redefined by the scaling of general-purpose large language models (LLMs) like GPT-3 \cite{brown2020language}, PaLM \cite{chowdhery2023palm}, and GPT-4 \cite{achiam2023gpt}, which demonstrated emergent capabilities in reasoning and generalization. This progression catalyzed a parallel evolution in the medical domain, transitioning from early BERT-based adaptations like ClinicalBERT \cite{huang2019clinicalbert} and BioBERT \cite{lee2020biobert} to expert-level generative systems. Recent milestones, such as MedPaLM \cite{singhal2023large} and Med-Gemini \cite{saab2024capabilities}, have successfully harnessed model scaling to achieve near-human performance on standardized medical examinations and complex clinical reasoning tasks.
To effectively adapt these powerful architectures for clinical utility, researchers employ specialized training and prompting strategies. Domain adaptation is primarily achieved through either training models from scratch on biomedical corpora, as exemplified by PubMedBERT \cite{gu2021domain} and BioGPT \cite{luo2022biogpt}, or through continuous pre-training of general checkpoints, as seen in ClinicalGPT \cite{wang2023clinicalgpt}. Furthermore, instruction tuning \cite{wei2021finetuned} has emerged as a critical alignment mechanism, where models are fine-tuned on expert-curated medical datasets to ensure safe and professional responses. Beyond training, inference-time strategies like Chain-of-Thought (CoT) prompting \cite{wei2022chain} have proven indispensable. By guiding models to decompose complex clinical scenarios into step-by-step reasoning paths, CoT aligns algorithmic processing with the structured diagnostic logic inherent to medical practice.

\subsection{Classical Reasoning Paradigms}

The paradigm of dividing reasoning into deductive reasoning, inductive reasoning, and abductive reasoning was first proposed by the renowned American philosopher Charles Sanders Peirce. This framework ingeniously utilizes the relationships among ``cases'', ``rules'', and ``results'' to clearly elucidate the essential differences between different types of reasoning. This classification framework has been employed in teaching reasoning-related fields for many years and has been widely adopted by the academic community.

\textbf{Deductive Reasoning} derives specific results from established general rules and known cases (Rule + Case $\rightarrow$ Result). It is characterized by \textit{irrefutability}: if the premises are true, the conclusion must necessarily be true \cite{Deductive_reasoning_2}. For instance, given the rule ``High fever indicates infection'' and the case ``This patient has a fever,'' the result ``This patient is infected'' is logically certain. While deduction ensures reliability in fields like mathematics, it does not generate new information beyond what is already implicit in the premises.

\begin{table}[!htbp]
\centering
\normalsize
\begin{tabular}{c|l l}
\hline
\multicolumn{3}{c}{\textbf{Deductive Reasoning}} \\
\hline
\multirow{2}{*}{\centering premises} & \textbf{Case} & This patient has a fever of 39°C.\\
 & \textbf{Rule} & High fever indicates infection. \\
\hline
conclusion & \textcolor{red}{\textbf{Result}} & This patient is infected. \\
\hline
\end{tabular}
\caption{Deductive Reasoning, Barbara Syllogism}
\label{tab:Deductive Reasoning}
\end{table}


\textbf{Inductive Reasoning} generalizes new rules from observed cases and results (Case + Result $\rightarrow$ Rule). Unlike deduction, its conclusions are \textit{defeasible} and probabilistic rather than certain \cite{Abductive_and_inductive_reasoning:_background_and_issues}. For example, observing that ``Many patients with high fever improved after antibiotics'' allows us to induce the general rule ``High fever indicates bacterial infection.'' Although this process can generate new knowledge necessary for scientific discovery, the conclusions remain subject to revision if counter-examples arise.

\begin{table}[!htbp]
\centering
\normalsize
\begin{tabular}{c|l l}
\hline
\multicolumn{3}{c}{\textbf{Inductive Reasoning}} \\
\hline
\multirow{2}{*}{\centering premises} & \textbf{Case} & Many patients have a fever of 39°C. \\
 & \textbf{Result} & They improved after antibiotics.\\
\hline
conclusion & \textcolor{red}{\textbf{Rule}} & High fever indicates infection. \\
\hline
\end{tabular}
\caption{Inductive Reasoning, Barbara Syllogism}
\label{tab:Inductive Reasoning}
\end{table}


\textbf{Abductive Reasoning} infers the most plausible case (cause) from known rules and observed results (Rule + Result $\rightarrow$ Case). It shares the defeasible nature of induction but focuses on explanation rather than generalization. In a clinical context, this corresponds to diagnostic reasoning: given the result ``The patient is infected'' and the rule ``High fever indicates infection,'' a physician abductively infers ``The patient likely has a fever'' as the explanation. This form of reasoning is fundamental to medical diagnosis, enabling physicians to hypothesize underlying causes from symptoms \cite{Abductive_and_inductive_reasoning:_background_and_issues}.

\begin{table}[!htbp]
\centering
\normalsize
\begin{tabular}{c|l l}
\hline
\multicolumn{3}{c}{\textbf{Abductive Reasoning}} \\
\hline
\multirow{2}{*}{\centering premises} & \textbf{Rule} &High fever indicates infection. \\
 & \textbf{Result} & This patient is infected.\\
\hline
conclusion & \textcolor{red}{\textbf{Case}} & This patient has a fever of 39°C. \\
\hline
\end{tabular}
\caption{Abductive Reasoning, Barbara Syllogism}
\label{tab:Abductive Reasoning}
\end{table}

After introducing the general concept of reasoning and its classical classification paradigms (deductive reasoning, inductive reasoning, and abductive reasoning), we can see that these reasoning theories not only constitute the core framework of human cognition and artificial intelligence, but are also widely applied and extended in specific domains. In the medical field, reasoning processes often involve complex clinical decision-making, diagnostic inference, and treatment planning, which essentially depend on the ability to derive conclusions from symptoms, test results, and medical knowledge. Applying general reasoning paradigms to medical contexts not only helps understand physicians' thought processes, but also provides a solid theoretical foundation for developing LLM medical reasoning taxonomy. Next, we will introduce the classical clinical competency grading framework to further understand the unique characteristics of medical reasoning.

\subsection{Levels of Clinical Competence}

To better understand the reasoning capabilities of Large Language Models in the medical domain, we establish a hierarchical classification of LLM reasoning from the perspective of medical competency grading. This classification not only helps define the capability boundaries of LLMs in medical reasoning, but also enables the assessment and regulation of their safety and effectiveness in actual clinical applications. Below, we provide a brief introduction to Miller's Pyramid \cite{The_assessment_of_clinical_skills/competence/performance}, the mainstream framework for medical competency grading that serves as background knowledge.

Miller’s Pyramid, introduced by George Miller in 1990, provides a foundational framework for understanding and assessing clinical competence. The pyramid consists of four levels: ``Knows'', ``Knows How'', ``Shows How'', and ``Does''. This framework has profound implications for understanding how LLMs might be integrated into medical practice and how their capabilities
should be evaluated.

\textbf{Knows} as the base of the pyramid, representing factual knowledge and information recall. At this level, assessment focuses on whether individuals can remember and recognize relevant medical facts, terminology, and concepts.

\textbf{Knows How} as the second Level of the pyramid, representing the ability to apply and interpret knowledge. At this level, the assessment focuses on whether individuals can understand and apply medical knowledge to solve problems, analyze situations, or formulate plans.

\textbf{Shows How} as the third Level of the pyramid, representing skill demonstration in simulated environments. At this level, the assessment focus is on whether individuals can demonstrate clinical skills and procedural operations under controlled conditions (such as laboratory or simulated scenarios).

\textbf{Does} as the top of the pyramid, representing authentic performance in real-world environments. At this level, the evaluation focuses on whether individuals can independently execute tasks, make decisions, and provide patient care in actual clinical practice.

Modern medical education and assessment have extended Miller's framework to address contemporary challenges in healthcare delivery and medical practice. These extensions are particularly relevant to understanding how LLMs might be integrated into clinical practice and what additional competencies they might need to develop.

\begin{figure}[!ht]
  \centering
  \includegraphics[width=0.99\linewidth, keepaspectratio]{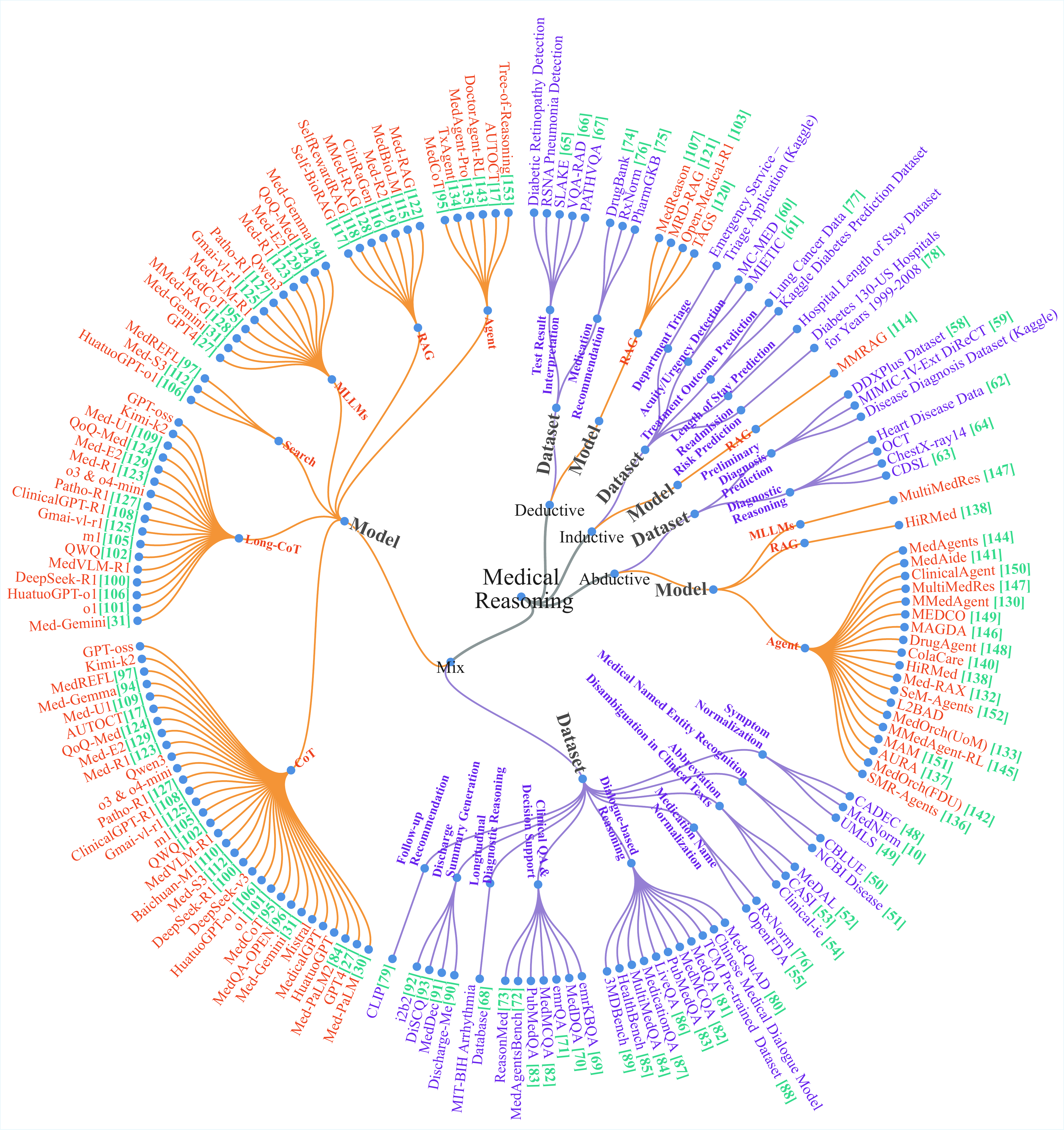}
  \caption{Overview of LLMs and datasets under different types of medical reasoning. The diagram organizes representative datasets and models according to four reasoning paradigms: deductive, inductive, abductive, and mixed. In addition, different types of LLMs are categorized, and datasets are further grouped by corresponding medical tasks.}
  \label{fig:reasoning-type}
\end{figure}

\subsection{Taxonomy of Medical Reasoning Capabilities}

Building upon the theoretical foundations of reasoning and clinical competence frameworks, we propose a comprehensive taxonomy for LLM medical reasoning capabilities. This taxonomy integrates classical reasoning paradigms with practical medical applications through two complementary organizational structures.

\subsubsection{Reasoning Type Classification}

Fig.~\ref{fig:reasoning-type} illustrates the mapping of medical tasks to four fundamental reasoning types. \textbf{Deductive reasoning tasks} involve applying established medical knowledge and guidelines to specific situations, such as symptom normalization and test result interpretation. \textbf{Inductive reasoning tasks} identify patterns in clinical data to make predictions, including length of stay prediction and readmission risk assessment. \textbf{Abductive reasoning tasks} generate and evaluate explanatory hypotheses for observed clinical phenomena, central to diagnostic processes. \textbf{Mixed reasoning tasks} combine multiple reasoning paradigms to address complex clinical scenarios requiring integrated decision-making.

\begin{figure*}[!ht]
  \centering
  \includegraphics[width=0.95\linewidth, keepaspectratio]{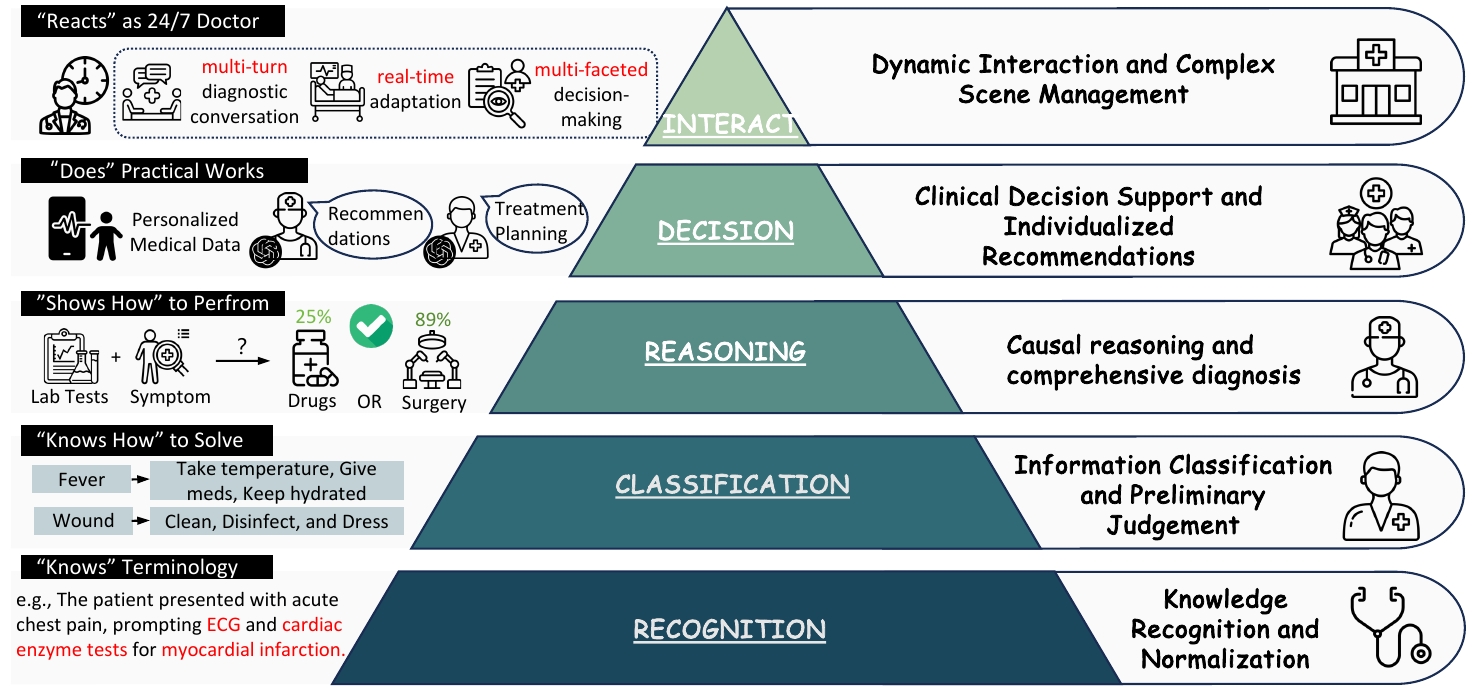}
  \caption{The proposed five-level competency hierarchy for evaluating LLM capabilities in medical reasoning. The hierarchy extends Miller’s Pyramid to the context of large language models, with ascending levels from basic medical knowledge recognition and normalization (Level 1) to dynamic interaction and complex scenario management (Level 5). Higher levels build upon competencies from the lower tiers, reflecting increasing reasoning sophistication and integration in clinical contexts.}
  \label{fig:medical-ability-level}
\end{figure*}

\subsubsection{Five-Level Competency Hierarchy}

Fig.~\ref{fig:medical-ability-level} presents our proposed five-level competency hierarchy that extends Miller's Pyramid to the context of large language models. This framework provides a structured approach to understanding and evaluating LLM capabilities in medical reasoning contexts, with each level corresponding to specific stages in Miller's classical framework for assessing clinical competence:

\textbf{Level 1: Medical Knowledge Recognition and Normalization (Knows)} forms the foundation, corresponding to Miller's ``Knows'' level. This involves accurate recognition, interpretation, and standardization of medical concepts and terminology. Systems must demonstrate ability to identify medical entities, understand semantic relationships, and map natural language to structured concepts in standardized medical knowledge bases. Typical tasks include symptom normalization, medical entity recognition, and medical terminology standardization.

\textbf{Level 2: Information Classification and Clinical Triage (Knows How)} aligns with Miller's ``Knows How'' level, involving categorization and prioritization of clinical information to support decision-making and resource allocation. This requires understanding of disease severity, urgency indicators, and institutional capabilities. Systems at this level can perform department triage prediction, urgency detection, and preliminary diagnostic classification, demonstrating knowledge of how medical principles should be applied.

\textbf{Level 3: Causal Reasoning and Comprehensive Diagnosis (Shows How)} corresponds to Miller's ``Shows How'' level, encompassing sophisticated diagnostic reasoning that integrates multiple information sources (including medical history, physical signs, laboratory results, and imaging) to reach comprehensive clinical conclusions. Systems must integrate diverse clinical information, perform causal reasoning, consider multiple diagnostic hypotheses, and handle uncertainty appropriately. This includes comprehensive diagnostic reasoning, test result interpretation, longitudinal diagnosis, and disease progression analysis.

\textbf{Level 4: Clinical Decision Support and Personalized Recommendations (Shows How/Does)} bridges Miller's ``Shows How'' and ``Does'' levels, focusing on treatment planning and clinical decision support. This requires integration of evidence-based medicine with individual patient factors and preferences to provide personalized recommendations. Systems at this level can perform medication recommendation, surgical/treatment planning, treatment outcome prediction, follow-up suggestions, and readmission risk assessment, demonstrating near-clinical performance capabilities.

\textbf{Level 5: Dynamic Interaction and Complex Scenario Management (Does)} represents the highest level of competence, corresponding to Miller's ``Does'' level. This encompasses dynamic, interactive reasoning in complex clinical scenarios that require real-time adaptation and multi-faceted decision-making. Systems must continuously collect information, adjust reasoning pathways, and manage complex cases through multi-turn diagnostic conversations, dynamic disease management, discharge summary generation, and comprehensive clinical question answering with decision support.

Each ascending level builds upon competencies from lower tiers, reflecting increasing reasoning sophistication and clinical integration. This hierarchical structure not only provides theoretical grounding through its alignment with Miller's Pyramid but also offers practical utility for researchers and practitioners in systematically evaluating and developing medical reasoning systems. The framework emphasizes the progression from foundational knowledge (``Knows'') through application understanding (``Knows How''), demonstrated competence (``Shows How''), to actual clinical performance (``Does''), mirroring the developmental trajectory expected in medical education and practice.

\section{Datasets and Benchmarks}

\subsection{General Workflow for Constructing Medical Datasets}
\begin{figure*}[!ht]
  \centering
  \includegraphics[width=0.99\linewidth, keepaspectratio]{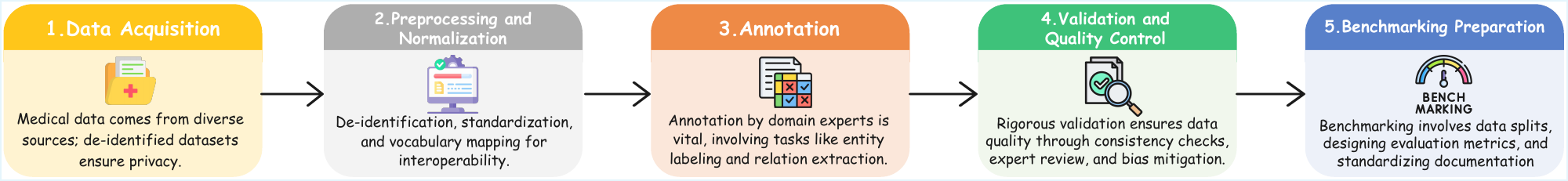}
  \caption{The general workflow for constructing medical datasets.}
  \label{fig:data-construction}
\end{figure*}

The construction of high-quality medical datasets is a prerequisite for advancing reasoning capabilities of large language models (LLMs) in healthcare. Unlike general-domain corpora, medical datasets must meet strict requirements regarding data authenticity, privacy protection, and domain specificity. A general workflow can be summarized in five stages: data acquisition, preprocessing and normalization, annotation, validation, and benchmarking preparation, which is shown in Fig.~\ref{fig:data-construction}.

\subsubsection{Data Acquisition}
Medical data can originate from diverse sources, including electronic health records (EHRs), clinical notes, diagnostic reports, biomedical literature, medical imaging, and patient-generated health data. To acquire such data typically requires collaboration with healthcare institutions and strict adherence to ethical and regulatory frameworks.

\subsubsection{Preprocessing and Normalization}
Raw medical data is often heterogeneous and noisy, containing free-text descriptions, structured tables, imaging formats, and laboratory values. Preprocessing involves de-identification of patient information, removal of artifacts, and normalization into standardized formats.  This step ensures interoperability and facilitates downstream reasoning tasks.

\subsubsection{Annotation}
Accurate and consistent annotation is crucial for supervised learning and evaluation. Annotation typically requires domain experts such as physicians or trained medical annotators. Depending on the task, annotation may involve entity labeling (e.g., symptoms, medications, diagnoses), relation extraction (e.g., drug–drug interactions), or reasoning-related tasks such as assigning diagnostic hypotheses or treatment recommendations. In this step, inter-annotator agreement measurement is essential for ensuring reliability.

\subsubsection{Validation and Quality Control}
Rigorous validation is necessary to guarantee dataset quality and usability. This involves internal consistency checks, cross-validation against external references (e.g., clinical guidelines), and expert review. Statistical analyses are often conducted to ensure representativeness across patient demographics, disease categories, and clinical settings. In addition, bias detection and mitigation are critical to support generalizable reasoning, such as addressing demographic imbalance or hospital-specific artifacts.

\subsubsection{Benchmarking Preparation}
Finally, datasets must be partitioned and standardized for benchmarking. Typical practices include dividing data into training, validation, and test sets; designing task-specific evaluation metrics; and defining challenge tasks aligned with reasoning paradigms. Public release requires careful documentation of dataset scope, limitations, and licensing terms. Increasingly, benchmark design also incorporates multi-institutional test sets and adversarial evaluation scenarios to assess robustness and fairness.

In summary, medical dataset construction is not merely a data collection exercise but a structured pipeline balancing clinical fidelity, ethical considerations, and technical requirements. Based on this workflow, reasoning-specific and competency-level benchmarks can be systematically developed.

\begin{table*}[!ht]
\footnotesize
\setlength{\tabcolsep}{5pt} 
\renewcommand{\arraystretch}{1.2}
\centering
\caption{Overview of datasets in different medical reasoning capabilities levels.}
\begin{threeparttable}
\resizebox{\textwidth}{!}{
\begin{tabular}{c|c|cccccc}
\toprule
\multirow{2}{*}{\large \textbf{\makecell{Capabilities}}} & 
\multirow{2}{*}{\large \textbf{Task}} & 
\multirow{2}{*}{\large \textbf{Dataset}} & 
\multirow{2}{*}{\large \textbf{Modality}} & 
\multirow{2}{*}{\large \textbf{Task Type}} & 
\multirow{2}{*}{\large \textbf{\makecell{Number of Samples}}} & 
\multirow{2}{*}{\large \textbf{Answer Form}} & 
\multirow{2}{*}{\large \textbf{Language}} \\
 & & & & & & & \\ \hline

\multirow{20}{*}{\textbf{{\fontsize{12pt}{14.4pt}\selectfont \makecell{Level1: \\ Knowledge \\ recognition \\ and \\ normalization}}}}
  & \multirow{6}{*}{{\fontsize{10pt}{12pt}\selectfont \makecell{Symptom \\ Normalization}}} & CADEC \cite{CADEC} & Txt & \makecell{NER, RE} & $\sim$1,250 posts & Free-Form & English \\ \cline{3-8}
  &  & UMLS \cite{UMLS} & Txt & \makecell{NER, RE} & - & Free-Form & English \\ \cline{3-8}
  &  & MedNorm \cite{MedNorm} & Txt & \makecell{NER, \\ Concept Normalization} & \makecell{27,979 \\ descriptions} & Free-Form & English \\ \cline{2-8}
  & \multirow{3}{*}{{\fontsize{10pt}{12pt}\selectfont \makecell{Medical Named \\ Entity Recognition} }}& CBLUE \cite{zhang-etal-2022-cblue} & Txt &  Multi-task NLU  & - & Free-form / labels & Chinese \\ \cline{3-8}
  &  & NCBI Disease \cite{NCBI} & Txt & \makecell{Disease NER \& \\ Concept Normalization}  & \makecell{793 abstracts } & \makecell{Span + \\ Concept ID }& English \\ \cline{2-8}
  & \multirow{5}{*}{{\fontsize{10pt}{12pt}\selectfont \makecell{Abbreviation \\ Disambiguation \\ in Clinical Texts} }} & MeDAL \cite{MeDAL} & Txt & Abbr. Disambig.  & \makecell{14,393,619 articles} & Open-text / label & English \\ \cline{3-8}
  &  & CASI \cite{CASI} & Txt & Abbr. Disambig.  & 440 abbreviations & Label & English \\ \cline{3-8}
  &  & Clinical-ie \cite{Clinical-ie} & Txt & \makecell{Information Extraction}  & \makecell{Depends on \\ subtasks} & \makecell{Span / \\ Structured output} & English \\ \cline{2-8}
  &{\fontsize{10pt}{12pt}\selectfont \makecell{Medication \\ Name Normalization}} & OpenFDA \cite{kass2016openfda} & Txt &  \makecell{Regulatory Retrieval}  & - & Structured JSON & English \\ \hline

\multirow{16}{*}{ \textbf{{\fontsize{12pt}{14.4pt}\selectfont \makecell{Level2: \\ Information \\ classification \\ and \\ preliminary \\ judgment}}}}
  &\multirow{6}{*}{{\fontsize{10pt}{12pt}\selectfont \makecell{Department \\ Triage} }}& MedDG \cite{liu2022meddg} & Txt & \makecell{Dialogue, \\ Response Generation }& \makecell{17,864 dialogues, \\ 385,951 utterances} & Open & Chinese \\ \cline{3-8}
  &  & CMID \cite{CMID} & Txt & \makecell{Intent Understanding}  & \makecell{$\sim$12K dialogues} & MC-QA & Chinese \\ \cline{3-8}
  &  & \makecell{ Emergency \\ Service–Triage Application\tnote{1}} & Tab & \makecell{Classification}  & $\sim$1267 instances  & Open & English \\ \cline{2-8}
  & \multirow{5}{*}{{\fontsize{10pt}{12pt}\selectfont \makecell{Preliminary \\ Diagnosis Prediction}}} & DDXPlus \cite{fansi2022ddxplus} & Tab &\makecell{Symptom Detection}  & $\sim$1.3M patients & Open & English \\ \cline{3-8}
  &  & MIMIC-IV-Ext DiReCT \cite{MIMIC-IV-Ext-DiReCT} & Txt &\makecell{Diagnostic Reasoning}  & 511 clinical notes & Open & English \\ \cline{3-8}
  &  & Disease Diagnosis\tnote{2} & Txt &\makecell{Diagnostic Reasoning}  & \makecell{$\sim$2,000 \\ instances} & Open & English \\ \cline{2-8}
  & \multirow{2}{*}{{\fontsize{10pt}{12pt}\selectfont \makecell{Acuity/Urgency \\ Detection}}} & MC-MED \cite{MC-MED} & Img,Txt & \makecell{Prediction} & \makecell{118,385 \\ visits} & Open & English \\ \cline{3-8}
  &  & \makecell{MIETIC}\cite{MIETIC} & Txt & \makecell{Triage Classification} & 9,629 triage cases  & Gen. & English \\ \hline

\multirow{20}{*}{ \textbf{{\fontsize{12pt}{14.4pt}\selectfont \makecell{Level3: \\ Causal \\ reasoning \\ and \\ comprehensive \\ diagnosis}}}}
  & \multirow{6}{*}{{\fontsize{10pt}{12pt}\selectfont \makecell{Diagnostic \\ Reasoning}}} & Heart Disease Data \cite{heart-disease-45} & Txt &\makecell{Prediction}  & 303 instances & Open & English \\ \cline{3-8}
  &  & CDSL \cite{CDSL} & Txt &\makecell{ Prediction }& 4,479 patients & Open & English \\ \cline{3-8}
  &  & OCT\tnote{3} & Img &\makecell{ Classification} & 84,495 images & Classification & English \\ \cline{3-8}
  &  & ChestX-ray14 \cite{ChestXray-NIHCC} & Img & \makecell{Disease Detection} & 112,120 images & Open & English \\ \cline{2-8}
  & \multirow{10}{*}{{\fontsize{10pt}{12pt}\selectfont \makecell{Test Result \\ Interpretation}}} & \makecell{ Diabetic \\  Retinopathy Detection\tnote{4}}& Img &\makecell{ Classification }& \makecell{$\sim$35,000 \\ images} & Classification & English \\ \cline{3-8}
  &  & \makecell{RSNA \\ Pneumonia Detection\tnote{5} }& Img &\makecell{ Detection} & \makecell{$\sim$30,000 \\ images} & Open & English \\ \cline{3-8}
  &  & SLAKE \cite{liu2021slake} & Img, Txt & \makecell{QA} & 14,028 QA pairs & Open & EN \& CN \\ \cline{3-8}
  &  & VQA-RAD \cite{lau2018dataset} & Img, Txt & \makecell{QA} & \makecell{315 images \\ 3.5K QA pairs} & Open & English \\ \cline{3-8}
  &  & PATHVQA \cite{he2020pathvqa} & Img, Txt & \makecell{QA} & \makecell{4,998 images \\ 32,799 QA pairs} & Open & English \\ \cline{2-8}
  &{\fontsize{10pt}{12pt}\selectfont \makecell{Longitudinal \\ Diagnostic Reasoning}} & MIT-BIH \cite{MIT-BIH-Arrhythmia-Database} & Txt &\makecell{Detection} & 48 instances & Classification & English \\ \hline

\multirow{30}{*}{\textbf{{\fontsize{12pt}{14.4pt}\selectfont \makecell{Level4: \\ Clinical \\ Decision \\ Support \\ and \\ Individualized \\ Recommendation}}}}
  & \multirow{10}{*}{{\fontsize{10pt}{12pt}\selectfont \makecell{Clinical QA \\ \& Decision Support} }}& emrKBQA \cite{emrKBQA} & Txt & \makecell{QA} & 940K question & Open & English \\ \cline{3-8}
  &  & MedDQA \cite{MedDQA} & Txt &\makecell{QA} & \makecell{2000 images} & MC-QA & Chinese \\ \cline{3-8}
  &  & emrQA \cite{emrQA} & Txt & \makecell{QA} & \makecell{1,957,835 QA pairs, \\ 1,225,369 QL pairs} & Open & English \\ \cline{3-8}
  &  & MedAgentsBench \cite{tang2025medagentsbench} & Txt & \makecell{QA} & \makecell{seven established \\ medical datasets} & Free-Form & English \\ \cline{3-8}
  &  & ReasonMed \cite{sun2025reasonmed} & Txt & \makecell{QA} & \makecell{370k examples} & Free-Form & English \\ \cline{2-8}
  & \multirow{5}{*}{{\fontsize{10pt}{12pt}\selectfont \makecell{Medication \\ Recommendation}}} & DrugBank \cite{DrugBank} & Txt & \makecell{Information Extraction} & \makecell{10,794 drugs, 1413413 \\ drug–drug  interactions} & Free-Form & English \\ \cline{3-8}
  &  & PharmGKB \cite{PharmGKB} & Txt &\makecell{Relation Extraction}  & - & Free-Form & English \\ \cline{3-8}
  &  & RxNorm \cite{RxNorm} & Txt &\makecell{ Drug Normalization} & \makecell{$>$100K  concepts} & Free-Form & English \\ \cline{2-8}
  & \multirow{3}{*}{{\fontsize{10pt}{12pt}\selectfont \makecell{Treatment \\ Outcome Prediction} }}& Lung Cancer Data \cite{lung_cancer_62} & Txt & \makecell{Survival Prediction }& \makecell{32 \\ instances} & Open & English \\ \cline{3-8}
  &  & \makecell{ Kaggle Diabetes \\ Prediction Dataset\tnote{6}}& Txt &\makecell{Prediction } & \makecell{768 \\ instances} & Open & English \\ \cline{2-8}
  &{\fontsize{10pt}{12pt}\selectfont \makecell{ Length of \\ Stay Prediction }}& \makecell{ Hospital Length \\ of Stay Dataset\tnote{7} }& Txt & \makecell{Prediction} & \makecell{$\sim$100K \\ instances} & Open & English \\ \cline{2-8}
  &{\fontsize{10pt}{12pt}\selectfont \makecell{ Readmission \\ Risk Prediction }}& \makecell{ Diabetes 130-US \\ Hospitals (1999–2008) \cite{diabetes_130-us_hospitals_for_years_1999-2008_296}} & Txt &\makecell{Classification}  & \makecell{101,766 \\ patient records} & Open & English \\ \cline{2-8}
  &{\fontsize{10pt}{12pt}\selectfont \makecell{Follow-up \\ Recommendation}} & CLIP \cite{CLIP} & Txt & \makecell{Multi-aspect extractive \\ summarization}  & 718 discharge & Free-Form & English \\ \hline

\multirow{30}{*}{\textbf{{\fontsize{12pt}{14.4pt}\selectfont \makecell{Level5: \\ Dynamic \\ Interaction \\ and \\ Complex \\ Scene \\ Management}}}}
  & \multirow{20}{*}{{\fontsize{10pt}{12pt}\selectfont \makecell{Dialogue-based \\ Reasoning}}} & Med-QuAD \cite{MedQuAD} & Txt & \makecell{QA} & $\sim$47K QA pairs  & Open & English \\ \cline{3-8}
  &  & MedQA \cite{MedQA} & Txt & \makecell{QA} & \makecell{61,097 questions} & MC-QA & EN \& CN \\ \cline{3-8}
  &  & MedMCQA \cite{MedMCQA} & Txt & \makecell{QA} & $\sim$194K questions & MC-QA & English \\ \cline{3-8}
  &  & PubMedQA \cite{PubMedQA} & Txt & \makecell{QA} & 211.3k QA pairs & Open & English \\ \cline{3-8}
  &  & MultiMedQA \cite{singhal2025toward} & Txt & \makecell{QA} & \makecell{$\sim$208k QA pairs} & MC-QA & English \\ \cline{3-8}
  &  & HealthBench \cite{arora2025healthbench} & Txt & \makecell{QA} & \makecell{5,000 conversations} & Free-Form & English \\ \cline{3-8}
  &  & LiveQA \cite{LiveMedQA2017} & Txt & \makecell{QA} & 634 QA pairs & Open & English \\ \cline{3-8}
  &  & MedicationQA \cite{MedicationQA} & Txt & \makecell{QA} & \makecell{674 QA pairs} & Open & English \\ \cline{3-8}
  &  &  \makecell{ Chinese Medical \\ Dialogue Model\tnote{8} }& Txt & \makecell{Dialogue, QA} & \makecell{$\sim$4K dialogs} & Open & Chinese \\ \cline{3-8}
  &  & TCM Pre-trained Dataset \cite{TCMChat} & Txt & \makecell{Pretraining} & \makecell{$\sim$100M  tokens} & Gen. & Chinese \\ \cline{3-8}
  &  & 3MDBench \cite{sviridov20253mdbench} & Img, Txt & \makecell{Dialogue} & 3013 cases & Free-Form & English \\ \cline{2-8}
  & \multirow{7}{*}{{\fontsize{10pt}{12pt}\selectfont \makecell{Discharge \\ Summary Generation}}} & Discharge-Me \cite{Discharge‑Me} & Txt & \makecell{Summarization} & \makecell{109,168  visits} & Gen. & English \\ \cline{3-8}
  &  & MedDec \cite{MedDec} & Txt & \makecell{Medical Decision \\ Extraction} & \makecell{451 discharge \\ summaries} & Free-Form & English \\ \cline{3-8}
  &  & \makecell{2012 i2b2 Temporal \\ Relations Dataset \cite{2012-i2b2-Temporal-Relations-Dataset} }& Txt &\makecell{Extraction}  & - & Free-Form & English \\ \cline{3-8}
  &  & DiSCQ \cite{DiSCQ} & Txt & \makecell{QA} & \makecell{2K+ questions} & Free-form & English \\ \bottomrule

\end{tabular}
}
\begin{tablenotes}
\footnotesize 
\item[1] \url{https://www.kaggle.com/datasets/ilkeryildiz/emergency-service-triage-application}
\item[2] \url{https://www.kaggle.com/datasets/s3programmer/disease-diagnosis-dataset} 
\item[3] \url{https://www.kaggle.com/datasets/paultimothymooney/kermany2018}
\item[4] \url{https://www.kaggle.com/c/diabetic-retinopathy-detection/data}
\item[5] \url{https://www.kaggle.com/c/rsna-pneumonia-detection-challenge/data}
\item[6] \url{https://www.kaggle.com/datasets/uciml/pima-indians-diabetes-database}
\item[7] \url{https://www.kaggle.com/datasets/aayushchou/hospital-length-of-stay-dataset-microsoft}
\item[8] \url{https://www.kaggle.com/datasets/thedevastator/chinese-medical-dialogue-model}
\end{tablenotes}
\end{threeparttable}
\end{table*}

\subsection{Datasets Categorized by Reasoning Types}
Medical reasoning tasks can be systematically organized according to the type of reasoning. Here we review datasets by four reasoning types: deductive, inductive, abductive, and mixed reasoning.

\subsubsection{Deductive Reasoning Datasets}
Deductive reasoning datasets emphasize the application of established medical knowledge, clinical guidelines, and standardized rules to specific patient cases. Unlike inductive or abductive paradigms, deductive reasoning is deterministic in nature, aiming to derive logically valid conclusions when the premises are correct. Accordingly, datasets in this category are often rule-grounded, designed to align with established medical ontologies and evidence-based protocols.

A defining characteristic of deductive reasoning datasets is their structured and standardized nature. The data typically includes explicit mappings between clinical inputs, such as symptom descriptions, laboratory results, or medication names. For example, patient-reported symptoms in free-text may need to be normalized to standardized vocabularies such as SNOMED CT or ICD codes, or laboratory values may be interpreted relative to reference ranges and diagnostic thresholds. These tasks demand precision and consistency, as small errors in normalization or interpretation can propagate downstream and compromise clinical decision-making.




Representative deductive reasoning datasets exemplify structured, rule-based alignment with clinical standards. For instance, the Diabetic Retinopathy Detection dataset\footnote{\url{https://www.kaggle.com/c/diabetic-retinopathy-detection/data}} provides retinal fundus images labeled with diabetic retinopathy severity, requiring models to interpret imaging findings against established diagnostic criteria. Similarly, the RSNA Pneumonia Detection dataset\footnote{\url{https://www.kaggle.com/c/rsna-pneumonia-detection-challenge/data}} offers annotated chest X-rays for pneumonia presence, demanding models to identify abnormalities and map them to clinical diagnoses.

\subsubsection{Inductive Reasoning Datasets}
Inductive reasoning datasets focus on identifying patterns in large-scale clinical data and generalizing these patterns to make probabilistic predictions about individual patients or populations. Unlike deductive tasks that rely on fixed guidelines, inductive datasets are inherently data-driven, capturing statistical regularities in patient outcomes, treatment responses, and disease trajectories.

Such datasets are usually derived from electronic health records (EHRs), intensive care unit (ICU) monitoring systems, or longitudinal patient databases. They often contain a mix of structured variables alongside unstructured clinical notes. The richness of these datasets enables predictive modeling, but also introduces challenges such as heterogeneity, imbalance, and institution-specific artifacts.


Representative inductive reasoning datasets capture statistical patterns in heterogeneous clinical data. For example, the Emergency Service – Triage Application dataset\footnote{\url{https://www.kaggle.com/datasets/ilkeryildiz/emergency-service-triage-application}} includes patient complaints, nursing assessments, triage levels, and department assignments, requiring models to predict triage severity.

\subsubsection{Abductive Reasoning Datasets}
Abductive reasoning datasets are designed to support the generation and evaluation of explanatory hypotheses for observed clinical phenomena. In contrast to deductive datasets and inductive datasets, abductive datasets aim to capture the diagnostic reasoning process, where multiple plausible explanations for patient presentations must be considered and weighed.

These datasets often involve complex, multi-source clinical information such as patient history, symptoms, imaging findings, and laboratory test results. They require models to reason under uncertainty and to prioritize hypotheses based on plausibility and supporting evidence. Unlike deductive tasks with definitive mappings, abductive datasets frequently admit multiple valid answers, reflecting the ambiguity inherent in real-world diagnosis.




Representative abductive reasoning datasets reflect diagnostic tasks that require generating and weighing multiple plausible explanations. The Heart Disease dataset (UCI)\cite{heart-disease-45} provides structured patient features such as blood pressure, ECG, and blood sugar, with outputs indicating disease presence, supporting inference of likely diagnoses from heterogeneous clinical signals. ChestX-ray14\cite{ChestXray-NIHCC} contains over 100,000 chest X-rays annotated for 14 thoracic conditions, requiring models to interpret imaging findings and propose probable disease explanations under uncertainty. 

\subsubsection{Mixed Reasoning Datasets}
Mixed reasoning datasets combine elements of deductive, inductive, and abductive reasoning, reflecting the hybrid and multi-step nature of real-world clinical practice.  A physician may begin by normalizing symptoms into standardized concepts (deductive), identify risk patterns from prior cases (inductive), and then generate plausible diagnostic hypotheses under uncertainty (abductive). Datasets in this category are therefore particularly valuable for evaluating whether LLMs can integrate multiple reasoning strategies in a coherent and clinically meaningful way.

Such datasets are usually broad in scope, spanning diverse task types such as medical question answering, clinical dialogue, and exam-style problem solving. They often require systems to first extract structured information, then apply probabilistic prediction, and finally reason about diagnostic or therapeutic options. 




Representative mixed reasoning datasets highlight the integration of deductive, inductive, and abductive processes. For example, CBLUE\cite{zhang-etal-2022-cblue}, a Chinese EMR entity recognition dataset, requires models to normalize clinical concepts using medical dictionaries (deductive) while generalizing entity patterns through statistical learning (inductive). NCBI Disease\cite{NCBI} provides PubMed abstracts annotated with disease entities mapped to standard ontologies, combining rule-based normalization with contextual pattern recognition.

\subsection{Datasets Categorized by Levels of Medical Competency}

To empirically ground the proposed five-level competency hierarchy, this section reviews existing datasets as practical benchmarks for each stage of medical reasoning. We systematically map available resources to the taxonomy, demonstrating how data characteristics evolve to match the increasing sophistication of the hierarchy: from foundational entity normalization (Level 1) and structured triage (Level 2), to complex causal diagnosis (Level 3) and personalized decision support (Level 4), culminating in dynamic, multi-turn interaction scenarios (Level 5). This organization explicitly connects theoretical competencies with their corresponding evaluation protocols, highlighting the specific data requirements necessary to assess AI capabilities at each stratum of the adapted Miller’s Pyramid.

\subsubsection{Level 1: Medical Knowledge Recognition and Normalization}

Datasets at this foundational level focus on the recognition, interpretation, and normalization of medical concepts from unstructured clinical or biomedical text. They correspond to the most basic reasoning capabilities, enabling systems to accurately identify mentions of diseases, symptoms, medications, and procedures, and to map them to standardized terminologies. This process ensures that heterogeneous clinical language can be reliably transformed into structured, interoperable knowledge. Precision in lexical and semantic mapping is critical, as errors at this stage may cascade and compromise higher-order reasoning tasks.

A specific feature of datasets at this level is their strong alignment with biomedical ontologies and controlled vocabularies. They are designed to test whether models can bridge free-text variability with standardized representations, such as SNOMED CT, ICD, or UMLS concepts. The emphasis lies not only on entity detection but also on disambiguation and normalization, ensuring consistency across different contexts of use.




Representative datasets at this level emphasize concept recognition and normalization. CADEC\cite{CADEC}, derived from medical forum posts, annotates drugs, adverse reactions, symptoms, and diseases, linking them to controlled vocabularies. MedNorm\cite{MedNorm} standardizes symptoms and adverse drug events from social media, requiring models to detect informal mentions and align them with UMLS concepts.

\subsubsection{Level 2: Information Classification and Clinical Triage}
Datasets at this level target the categorization, prioritization, and contextual interpretation of clinical information to support decision-making and resource allocation. Unlike  datasets in level 1 that focus on concept recognition, Level 2 datasets assess whether systems can apply medical knowledge to evaluate the severity, urgency, or relevance of patient information. This requires models to interpret clinical notes, laboratory results, or symptom reports in context and to assign appropriate classifications such as triage levels, department referrals, or risk strata.

The feature of Level 2 datasets is their focus on structured decision labels derived from expert judgments or established protocols. The data is often organized to facilitate evaluation of classification performance across multiple clinical dimensions, including urgency (e.g., emergent vs. non-emergent), specialty assignment (e.g., cardiology vs. gastroenterology), and risk stratification (e.g., high vs. low readmission probability). These datasets provide benchmarks for testing whether LLMs can move beyond entity recognition to context-aware reasoning, integrating multiple pieces of information to reach actionable conclusions.

Representative datasets for level 2 evaluate context-aware classification and preliminary diagnosis. DDXPlus\cite{fansi2022ddxplus} maps structured symptom descriptions to multiple possible diagnoses. MIMIC-IV-Ext DiReCT\cite{MIMIC-IV-Ext-DiReCT} extends MIMIC-IV with clinician-annotated SOAP notes and reasoning trees, challenging models to reconstruct likely discharge diagnoses from incomplete patient records. The Disease Diagnosis Dataset\footnote{\url{https://www.kaggle.com/datasets/s3programmer/disease-diagnosis-dataset}}
 provides simulated patient cases with demographics, symptoms, vital signs, and annotated diagnoses with severity, testing models’ ability to classify both disease type and urgency.

\subsubsection{Level 3: Causal Reasoning and Comprehensive Diagnosis}
Datasets at this level emphasize the integration of diverse clinical information to support comprehensive diagnostic reasoning. Unlike Level 1 and Level 2 datasets, which focus on concept recognition or classification, Level 3 datasets require systems to consider multiple data sources, such as patient history, laboratory tests, imaging reports, and symptom trajectories, and generate well-reasoned conclusions about potential diagnoses. This involves causal inference, temporal reasoning, and evaluation of competing hypotheses under uncertainty, reflecting the complexity of real-world clinical decision-making.

A defining characteristic of Level 3 datasets is their multi-modal and multi-step nature. Cases are often structured as complex clinical scenarios or narratives, where correct interpretation depends on understanding relationships between findings, temporal progression of symptoms, and underlying pathophysiology. Annotations frequently include primary diagnoses, differential diagnoses, and relevant supporting evidence, providing rich supervision for evaluating system reasoning and explanatory capabilities.



Representative Level 3 datasets highlight multi-modal integration and complex diagnostic reasoning. OCT (Optical Coherence Tomography)\footnote{\url{https://www.kaggle.com/datasets/paultimothymooney/kermany2018}} provides large-scale retinal images for detecting diseases such as macular degeneration, diabetic retinopathy, and glaucoma, requiring models to infer plausible diagnoses from imaging data under uncertainty. CDSL (COVID-19 Data Shared Learning)\cite{CDSL} combines vital signs, laboratory tests, and imaging data from COVID-19 patients, demanding integration of heterogeneous inputs to determine infection status and severity.

\subsubsection{Level 4: Clinical Decision Support and Personalized Recommendations}
Datasets at this level focus on treatment planning and clinical decision support, requiring systems to integrate evidence-based guidelines with individual patient characteristics. Unlike earlier levels, which emphasize recognition, classification, or diagnostic reasoning, Level 4 datasets test whether models can recommend personalized interventions, account for contraindications, and predict treatment outcomes. These datasets are crucial for evaluating a system’s ability to move from diagnostic insight to actionable clinical guidance.

In the datasets in level 4, they combine the patient-specific features with structured clinical knowledge. The data often includes demographics, comorbidities, laboratory and imaging results, and historical treatment responses, linked to recommended therapeutic options and expected outcomes. This allows evaluation of whether systems can not only select guideline-concordant treatments but also tailor recommendations to individual patient contexts, reflecting real-world complexity in clinical decision-making.

Representative Level 4 datasets emphasize treatment planning and clinical decision support through mixed reasoning. emrKBQA\cite{emrKBQA} links physician questions, logical forms, and structured patient records, requiring integration of knowledge with patient-specific data to generate accurate recommendations. 
Similarly, EHRSQL \cite{lee2022ehrsql} presents a text-to-SQL benchmark for question answering on electronic health records, requiring precise structured reasoning.
MedMCQA\cite{MedMCQA} presents large-scale multiple-choice medical questions, testing both deductive knowledge recall and abductive contextual inference.

\subsubsection{Level 5: Dynamic Interaction and Complex Scenario Management}
Datasets at this highest competency level evaluate dynamic, interactive reasoning in complex clinical scenarios that involve evolving patient conditions and multi-faceted decision-making. Unlike Level 4 datasets, which focus on static treatment planning, Level 5 datasets test whether systems can adapt reasoning strategies in real time, coordinate multiple aspects of care, and manage multi-system clinical interactions. This level represents the most sophisticated form of medical reasoning, closely resembling the full spectrum of clinical practice.

The defining feature of Level 5 datasets is their emphasis on multi-turn tasks. Cases often simulate real-world clinical workflows, including diagnostic interviews, care coordination among multiple providers, and longitudinal monitoring of patient responses. Annotated outcomes may include multi-step decisions, patient-centered care adjustments, and dynamic prioritization under changing conditions.

Level 5 datasets test dynamic, interactive reasoning in clinical scenarios. 
Med-QuAD~\cite{MedQuAD} provides multi-turn medical dialogues for interpreting evolving information and inferring potential causes, while the Chinese Medical Dialogue Model\footnote{\url{https://www.kaggle.com/datasets/thedevastator/chinese-medical-dialogue-model}} simulates patient–doctor conversations, requiring iterative information gathering and reasoning refinement.

\section{Paradigms of Medical LLM Reasoning}

\begin{figure*}
    \centering
    \includegraphics[width=\linewidth]{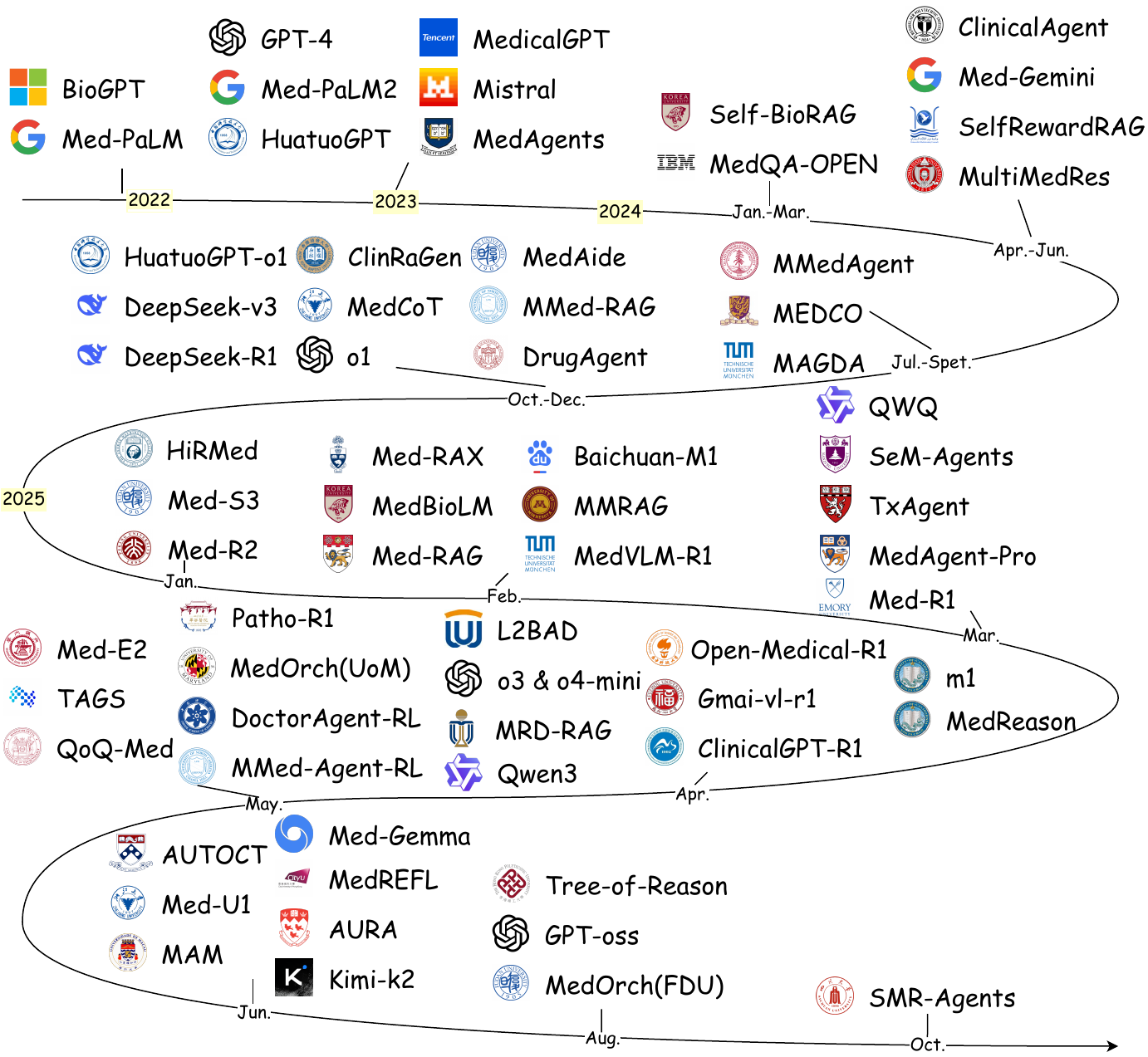}
    \caption{An evolutionary timeline illustrating the progression of LLM-based methods in medical AI.}
    \label{fig:tree-timeline}
\end{figure*}

\begin{figure*}
    \centering
    \includegraphics[width=\linewidth]{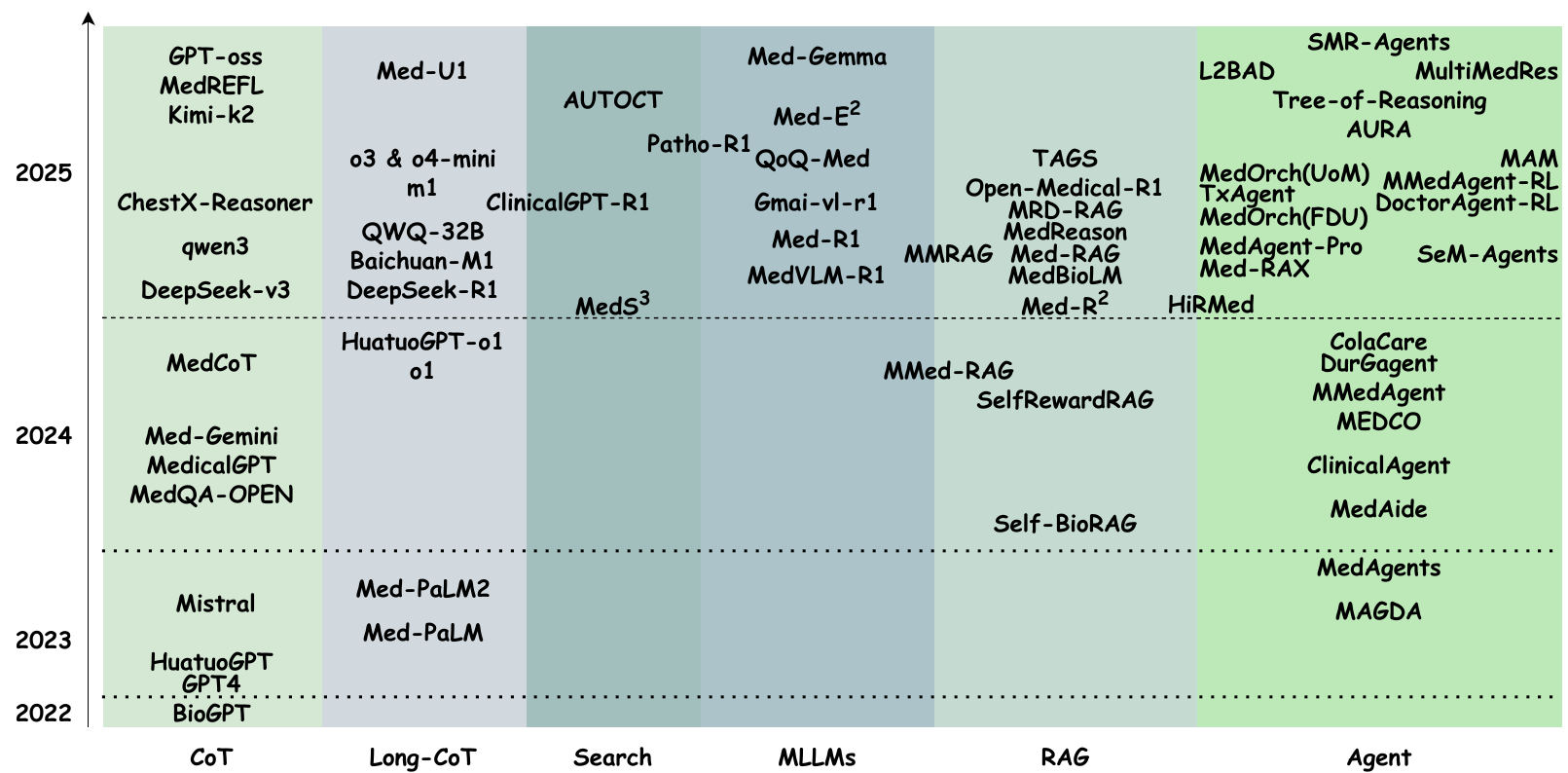}
    \caption{Different methodological categories of medical LLMs.}
    \label{fig:model-cate}
\end{figure*}

\begin{figure*}
    \centering
    \includegraphics[width=\linewidth]{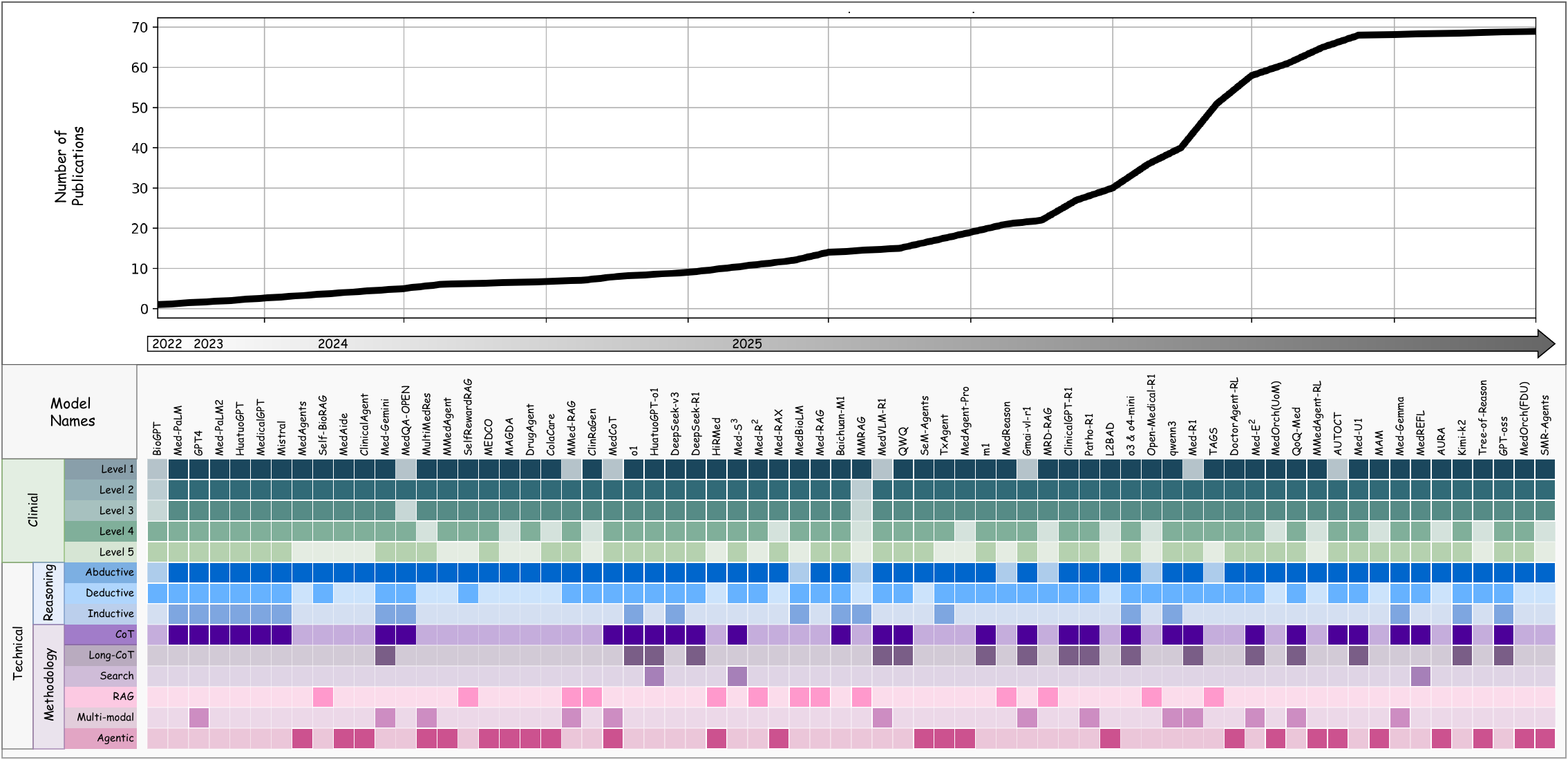}
    \caption{The evolution history of medical LLM reasoning.}
    \label{fig:his}
\end{figure*}

\begin{figure*}
    \centering
    \includegraphics[width=0.9\linewidth]{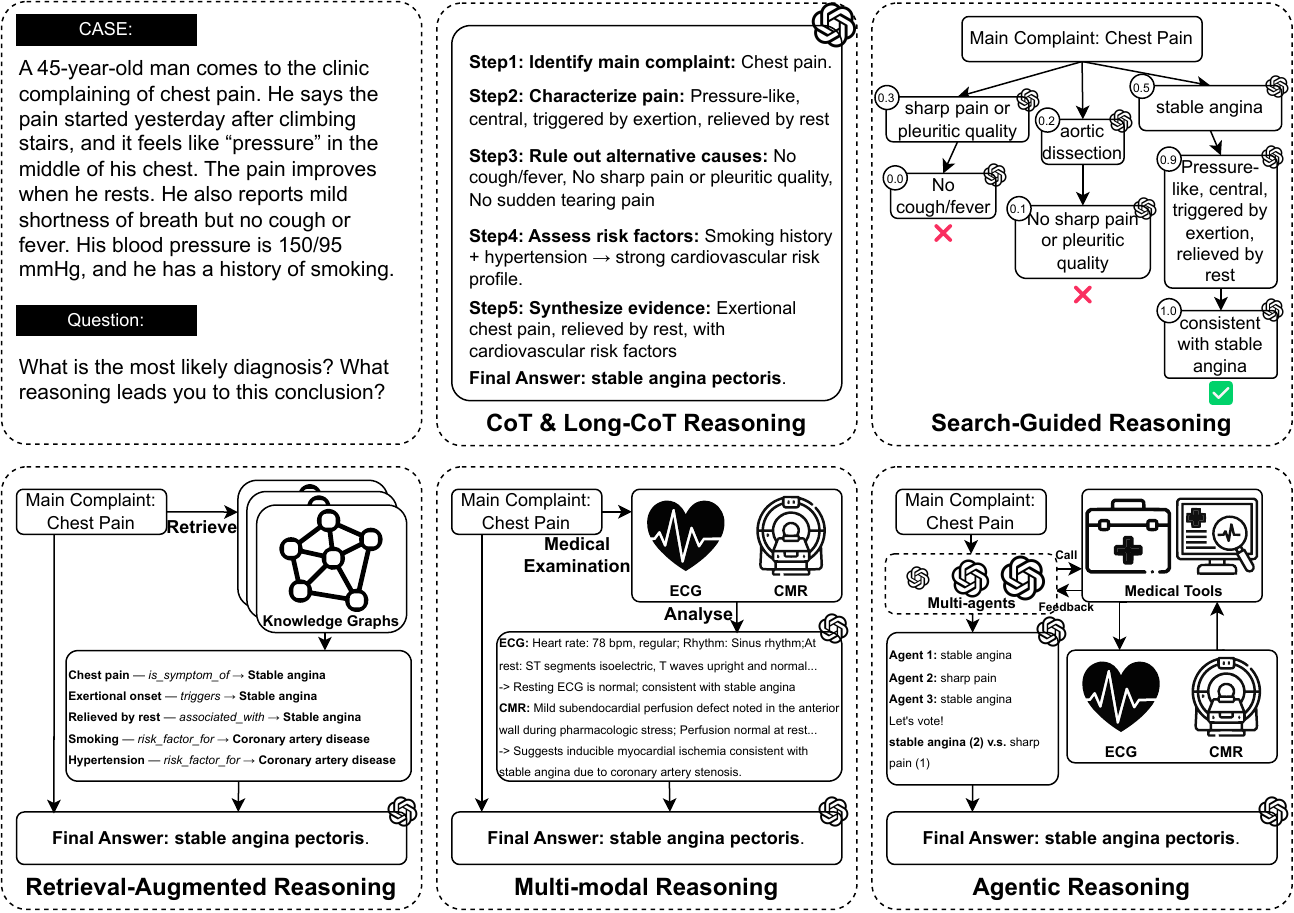}
    \caption{The comparison of various medical LLM Reasoning paradigms.}
    \label{fig:rlms}
\end{figure*}

Recent advances in Large Language Model (LLM) reasoning have shown remarkable potential in the medical domain, where tasks often require complex, multi-step, and explainable reasoning. In this section, we first present the evolutionary timeline of medical LLMs in Fig.~\ref{fig:tree-timeline}, followed by an overview of the methodological categories of representative general-purpose and medical LLMs in Fig.~\ref{fig:model-cate}. We then review the major reasoning paradigms employed in medical LLMs, as illustrated in Fig.~\ref{fig:his} and~\ref{fig:rlms}, including Chain-of-Thought reasoning, Long-CoT reasoning, search-guided reasoning, retrieval-augmented reasoning, and multimodal reasoning.
Fig.~\ref{fig:his} provides a comprehensive mapping of representative medical LLMs against our proposed five-level competency hierarchy and their underlying methodologies. The heatmap reveals a clear distributional trend that mirrors the hierarchy's increasing complexity: while the majority of surveyed models demonstrate robust proficiency in foundational tasks (Levels 1 and 2), represented by dense coverage in the upper rows, capability saturation drops significantly at the advanced stages (Levels 4 and 5). Crucially, the figure makes the connection between competency and methodology explicit; models achieving high-level proficiency (e.g., Level 5: Dynamic Interaction) exhibit a strong correlation with advanced reasoning paradigms such as Long-CoT and Agentic workflows, whereas models limited to lower levels predominantly rely on standard Chain-of-Thought or basic retrieval strategies. This visual evidence underscores that ascending the competency hierarchy requires a commensurate evolution in algorithmic sophistication.

\subsection{Chain-of-Thought Reasoning}

Chain-of-Thought (CoT) \cite{wei2022chain} reasoning can be formalized as a process where an LLM generates a sequence of linear reasoning steps, $R = (r_1, r_2, \dots, r_n)$, which are interleaved with the final answer, $A$. The entire process is usually a single, continuous generation prompted by ``\texttt{Let's think step by step}".

The output, which includes both the reasoning steps and the final answer, can be represented as a single output stream produced by a model $\mathcal{M}$ on a given input $\mathcal{I}$. The formalization can be expressed as:
\begin{equation}
(r_1, r_2, \dots, r_n, A) = \mathcal{M}(\mathcal{I}).
\end{equation}

In this framework, each reasoning step $r_i$ and the final answer $A$ are generated sequentially, while each step is entirely dependent on the previous steps, which can be viewed as an auto-regressive process:
\begin{equation}
\begin{aligned}
r_1 &= \mathcal{M}(\mathcal{I}) \\
r_2 &= \mathcal{M}(\mathcal{I}, r_1) \\
&\vdots \\
A &= \mathcal{M}(\mathcal{I}, r_1, r_2, \dots, r_n).
\end{aligned}
\end{equation}

In the medical domain, LLMs, such as GPT-4 \cite{achiam2023gpt}, Med-PaLM \cite{singhal2023large}, Med-PaLM 2 \cite{singhal2025toward}, Med-Gemini \cite{saab2024capabilities}, MedGemma \cite{sellergren2025medgemma} and etc., all demonstrate powerful CoT reasoning capabilities, generalizing well across various medical tasks.
However, beyond these LLMs with inherent CoT features, there are also several works trying to enhance their CoT reasoning ability.
For instance, MedCoT \cite{liu2024medcot} proposes a hierarchical, multi-expert CoT approach for Medical VQA, where an initial specialist produces a reasoning chain, a follow-up specialist validates it.
MEDQA-OPEN \cite{nachane2024few} introduces a modified open-ended version of MedQA with clinician-verified reasoning answers, and proposes a novel few-shot CoT prompting scheme that mimics incremental clinical reasoning.
Besides, Med-REFL \cite{yang2025med} constructs preference CoT pairs via the Tree-of-Thought \cite{yao2023tree} prompting strategy and applies Direct Preference Optimization \cite{rafailov2023direct} to train the reasoning capability of the LLM.

\subsection{Long Chain-of-Thought Reasoning}

Building upon the Chain-of-Thought reasoning, Long Chain-of-Thought (Long-CoT) reasoning takes a step further by encouraging or enabling the model to explore the problem space via significantly more extensive and detailed reasoning processes \cite{guo2025deepseek}. While standard CoT might involve a few concise linear exploration of the problem space, Long-CoT is characterized by its depth, verbosity, and the inclusion of sub-steps, explorations of alternative paths, self-correction, and explicit verification \cite{chen2025towards}.

The formalization of Long-CoT retains a similar structure as standard CoT, but with an emphasis on the properties of the reasoning processes $R_{long} = (r_1, r_2, \dots, r_m)$, where $m \gg n$. Generally, Long-CoT reasoning differs previous CoT in the following aspects:
\begin{itemize}
    \item \textbf{Increased Length and Granularity}: The number of reasoning steps is substantially larger. Each step $r_i$ might be more granular, breaking down a single logical leap from standard CoT into multiple, smaller, and more explicit deductions.
    \item \textbf{Structural Complexity}: The reasoning chain $R_{long}$ is not strictly linear. It can include branching, self-correction, and verification processes that help recursively explore the problem space.
    \item \textbf{Enhanced Information Synthesis}: The longer process allows the model to more thoroughly synthesize information from the input $\mathcal{I}$ connect disparate concepts, and manage a larger working memory of intermediate facts and sub-problems.
\end{itemize}

Owing to enhanced granularity in exploring the problem space, Long-CoT reasoning LLMs, such as OpenAI o1~\cite{jaech2024openai}, DeepSeek-R1~\cite{guo2025deepseek}, and QWQ-32B~\cite{qwq32b}, have demonstrated the effectiveness of Long-CoT reasoning in solving complex commonsense and mathematical problems. Notably, DeepSeek-R1 firstly leverages a two stage training pipeline, combining Cold-start Supervised Fine-tuning (SFT) with Group Relative Policy Optimization (GRPO), a reinforcement learning (RL) framework to encourage deeper, multi-step reasoning paths, thereby enhancing both accuracy and interpretability.

In the medical domain, the Long-CoT reasoning paradigm has catalyzed the development of specialized LLMs that prioritize interpretability and diagnostic precision, generating detailed reasoning chains before producing final predictions, thereby enhancing transparency and clinical trust. Generally, there are two main directions in applying the DeepSeek-R1 strategy to train these medical Long-CoT reasoning models:

\noindent\textbf{1) create high-quality reasoning paths.}
The quality of reasoning paths is paramount, as they determine the model's ability to move beyond simple pattern recognition to provide transparent, trustworthy, and evidence-based conclusions.

Specifically, Open-Medical-R1 \cite{qiu2025open}, built on the Open-R1 \cite{openr1} framework, explores optimal data distillation strategies to construct the reasoning dataset.
Similarly, m1 \cite{huang2025m1} proposes a novel data curation pipeline, applying Difficulty Filtering, Thinking Generation, and Diversity Sampling to select a 1K high-quality subset from the original Medical QA Pairs.
HuatuoGPT-o1 \cite{chen2024huatuogpt} pioneers the concept of ``verifiable medical questions,” integrating search-based reasoning to construct clinically grounded CoT paths,
while MedReason \cite{wu2025medreason} transforms QA pairs into logical paths using structured knowledge graphs, where each path is validated for consistency with clinical logic and evidence-based medicine. 
Furthermore, ClinicalGPT-R1 \cite{lan2025clinicalgpt} combines 20,000 real-world records with synthetic data, leveraging state-of-the-art LLMs combined with a long-chain reasoning prompt strategy to generate both steps and final results.
In parallel, EHR-R1 \cite{liao2025ehr} enhances the reasoning capabilities of foundational models specifically for electronic health record analysis by integrating chain-of-thought fine-tuning.

\noindent\textbf{2) optimize the RL framework.}
By optimizing the RL framework, researchers can train models to better generalize across diverse tasks and penalize flawed reasoning, ultimately pushing the model's capabilities from a knowledge retrieval system toward an expert-level, decision-making assistant.

For instance, Med-U1 \cite{zhang2025med} proposes a unified RL framework for diverse QA formats, optimizing multi-objective rewards and penalizing shallow reasoning chains, resulting in strong generalization across out-of-distribution tasks. 
Similarly, Baichuan-m1 \cite{wang2025baichuan} leverages a three-stage alignment strategy: Exploratory Log-likelihood Optimization (ELO), Token-Level Direct Preference Optimization (TDPO), and Proximal Policy Optimization (PPO), on top of a massive multilingual corpus and a proprietary medical knowledge base, enabling expert-level evidence-based reasoning.

\subsection{Search-Guided Reasoning}

Search-guided reasoning refers to methods where the reasoning process of LLMs is actively directed by a structured search algorithm, such as Monte Carlo Tree Search (MCTS) \cite{browne2012survey, zhang2025llama}. Instead of greedily following a single reasoning chain, the model explores multiple candidate reasoning trajectories, guided by exploration–exploitation trade-offs.  
The MCTS procedure consists of four phases: selection, expansion, simulation, and backpropagation. In the selection step, the next action $a^*$ is chosen according to the Upper Confidence Bound (UCB) criterion:
\begin{equation}
    a^* = \arg\max_a \; Q(s,a) + c \cdot \sqrt{\frac{\ln N(s)}{N(s,a)}},
\end{equation}
where $Q(s,a)$ is the estimated value of taking action $a$ in state $s$, $N(s)$ is the number of visits to state $s$, $N(s,a)$ is the number of visits to the $(s,a)$ pair, and $c$ is a tunable exploration parameter.

Therefore, given an LLM $\mathcal{M}$, the input $\mathcal{I}$, and the current state $s_t=(r_1,r_2, ...,r_t)$, which records the reasoning paths, the LLM will explore the problem space and generate $K$ viable reasoning paths (or actions):
\begin{equation}
    \{r^1_{t+1},r^2_{t+1}, ..., r^K_{t+1}\} = \mathcal{M}(\mathcal{I},s_t)
\end{equation}
Next, from the set of candidate steps generated by the LLM, $r^*_{t+1}$ is selected by the UCB criterion:
\begin{equation}
    r^*_{t+1} = \underset{r\in\{r^1_{t+1},r^2_{t+1}, ..., r^K_{t+1}\}}{\arg\max} \; Q(s_t,r) + c \cdot \sqrt{\frac{\ln N(s_t)}{N(s_t,r)}},
\end{equation}
Finally, once the best step $r^*_{t+1}$ is selected, the state (or the reasoning paths) is updated by appending this new step. This represents the expansion of the reasoning path:
\begin{equation}
    s_{t+1} = (r_1,r_2, ...,r_t,r^*_{t+1}).
\end{equation}

In the context of medical reasoning, search-guided approaches enable the model to evaluate multiple diagnostic hypotheses, simulate different treatment strategies, and select the most promising reasoning path based on accumulated evidence. Specifically, MedS$^3$ \cite{jiang2025meds} introduces a novel reasoning evolution framework using MCTS and Process Reward Models (PRM), along with a Vote-Sum strategy for inference-time selection, achieving significant improvement over 32B generalist models with a smaller scale. Similarly, AUTOCT \cite{liu2025autoct} proposes a multi-agent framework that applies MCTS to iteratively optimize clinical trial prediction performance.

\subsection{Retrieval-Augmented Reasoning}

Retrieval-Augmented Reasoning combines LLM inference with an external retrieval component to inject up-to-date, domain-specific information \cite{arslan2024survey}. In medical domain, this approach addresses the challenge of outdated pretraining data by dynamically fetching relevant guidelines, research literature, and patient-specific records during reasoning \cite{bora2024systematic}.  
Retrieval-Augmented Reasoning can be decomposed into three stages: retrieval, fusion, and reasoning. 

Formally, given an input query $\mathcal{I}$ and a medical knowledge base $D$, the retrieval stage selects a subset of relevant documents:
\begin{equation}
    R = \text{Retriever}(\mathcal{I}, D),
\end{equation}
where $R = \{d_1, d_2, \ldots, d_k\}$ denotes the retrieved documents. 

In the fusion stage, the retrieved documents are encoded and aggregated into a contextual representation:
\begin{equation}
    C_D = \mathcal{F}_{\text{fusion}}(\mathcal{I}, R) 
      = \sum_{i=1}^{k} w_i \cdot \phi(d_i),
\end{equation}
where $\phi(d_i)$ is the representation of document $d_i$ and $w_i$ is its relevance weight (e.g., based on similarity scores or domain-specific priors). 
Finally, the reasoning stage generates the answer by conditioning on both the input query and the fused representation:
\begin{equation}
    A = \mathcal{M}_{RAG}(\mathcal{I}, C_D).
\end{equation}

Beyond textual documents, a medical knowledge graph $G = (V, E)$ can also serve as an external knowledge source. 
Given a query $\mathcal{I}$, the retriever extracts a subgraph $G_{\mathcal{I}} \subseteq G$ containing relevant entities and relations. 
The subgraph is encoded into a representation by aggregating over its nodes and edges:
\begin{equation}
   C_{KG} = \mathcal{F}_{\text{graph}}(G_{\mathcal{I}}) 
          = \sum_{(u,v) \in E_{\mathcal{I}}} \alpha_{uv} \cdot \psi(u,v),
\end{equation}
where $\psi(u,v)$ encodes the relation between entities $u$ and $v$, and $\alpha_{uv}$ is an attention weight reflecting the importance of this relation. 

In this case, the reasoning stage generates the final answer based on both the input query and the fused representation of the retrieved subgraph:
\begin{equation}
   A = \mathcal{M}_{KG-RAG}(\mathcal{I}, C_{KG}).
\end{equation}

Existing medical RAG frameworks mainly focus on the refinement of the retrieval quality for medical documents. For instance, MMRAG \cite{zhan2025mmrag} retrieves similar examples and lets LLMs to reason in context for biomedical information extraction, while MedBioLM \cite{kim2025medbiolm} designs a knowledge searching and retrieving stage to acquire relevant knowledge, thereby improving the reasoning abilities of LLMs and the factual accuracy. Similarly, ClinRaGen \cite{niu2024multimodal} also proposes to retrieve disease-related documents from the knowledge base to combine LLM-driven reasoning with knowledge augmentation.
Building upon this, Self-BioRAG \cite{jeong2024improving} and SelfRewardRAG \cite{hammane2024selfrewardrag} further retrieves domain-specific documents, and applies self-reflection on generated responses to refine the retrieval results.
Moreover, Med-R$^2$ \cite{lu2025med} introduces the Evidence-Based Medicine (EBM) process to improve the trustworthiness of the retrieval stage, while TAGS \cite{wu2025tags} proposes a hierarchical retrieval mechanism that selects similar context examples based on both semantic and rationale-level.

Meanwhile, models such as MRD-RAG \cite{chen2025mrd} and MedRAG \cite{zhao2025medrag} are trying to incorporate comprehensive knowledge graphs as the knowledge base. The incorporation of structural knowledge allows these models to leverage the relationships and entities within the graph to provide more nuanced and accurate responses. 

\subsection{Multimodal Reasoning}

Medical decision-making often involves integrating heterogeneous data modalities such as clinical text and medical images \cite{boehm2022harnessing}. 
Multimodal reasoning in LLMs aims to jointly process and reason over such inputs to produce coherent outputs. 

Let $\mathcal{I}_{\text{text}}$ and $\mathcal{I}_{\text{img}}$ denote the textual and imaging inputs respectively. 
First, modality-specific encoders are used to obtain their representations:
\begin{equation}
   h_{\text{text}} = \phi_{\text{text}}(\mathcal{I}_{\text{text}}), \quad
   h_{\text{img}}  = \phi_{\text{img}}(\mathcal{I}_{\text{img}}).
\end{equation}
These representations are then integrated through a multimodal fusion function:
\begin{equation}
   C = \mathcal{F}_{\text{fusion}}(h_{\text{text}}, h_{\text{img}})
     = \sum_{m \in \{\text{text}, \text{img}\}} \alpha_m \cdot h_m,
\end{equation}
where $\alpha_m$ denotes the modality-specific attention weight. 
Finally, the multimodal reasoning language model generates the output:
\begin{equation}
   A = \mathcal{M}_{mm}(\mathcal{I}_{\text{text}}, \mathcal{I}_{\text{img}}, C).
\end{equation}

In multi-modal medical tasks, several recent works have introduced innovative strategies to enhance the reasoning capabilities of large vision-language models (LVLMs). These LVLMs are required to comprehend and perform reasoning on these medical images to provide accurate diagnosis and predictions.

Specifically, Med-R1 \cite{lai2025med}, QoQ-Med \cite{dai2025qoq}, and Gmai-vl-r1 \cite{su2025gmai} apply the DeepSeek-R1 strategy in LVLMs, enabling these models to support diverse medical imaging modalities, such as CT, X-ray, and MRI, across a variety of diagnostic and analytical tasks, aiming to create versatile tools for clinical applications. 

Meanwhile, there are also works focused on specifical modalities. 
For example, ChestX-Reasoner \cite{fan2025chestx} enhances reasoning by implementing a step-by-step verification process, which is particularly effective for radiology foundation models,
Similarly, Patho-R1 \cite{zhang2025patho} focuses on multimodal pathology reasoning, employing Group Relative Policy Optimization and Decoupled CLIP Sampling across 3.5M image-text pairs and 500K CoT samples.

Besides, some works are trying to improving the logical reasoning and factual grounding or develop more efficient training pipelines for medical LVLMs.
Here, MMed-RAG \cite{xia2024mmed} introduces a domain-aware retrieval mechanism that retrieves multi-modal documents during the fine-tuning stage.
At the same time, Med-E$^2$ \cite{mu2025elicit} proposes a two-stage post-training pipeline that elicits and then enhances multimodal reasoning for medical domains, providing a structured and systematic way to improve model performance in medical domains.

\subsection{Agentic Reasoning}
Agentic reasoning in medicine extends beyond passive information retrieval by enabling LLMs to actively plan, decide, and interact with external tools or environments during the reasoning process \cite{li2024mmedagent, schmidgall2024agentclinic}. 
Unlike traditional RAG systems that primarily focus on enriching the input with domain-specific knowledge, agentic frameworks introduce autonomy by allowing LLMs to decompose complex clinical queries, orchestrate multi-step actions, and verify their outputs. 
This paradigm is particularly critical in medical contexts where diagnostic reasoning, treatment planning, and patient-specific recommendations require both factual correctness and adaptive decision-making.

Formally, the process can be decomposed into planning, action, interaction, and verification stages. 
Given a medical query $\mathcal{I}$, the planning module generates a sequence of reasoning steps:
\begin{equation}
   \pi = \mathcal{P}(\mathcal{I}) = \{s_1, s_2, \ldots, s_T\},
\end{equation}
where $s_t$ denotes the $t$-th subtask or reasoning step. 
At each step, the agent selects an action based on the current state and available external tools:
\begin{equation}
   a_t = \pi(s_t, \mathcal{E}),
\end{equation}
where $\mathcal{E}$ represents the set of external tools or environments (e.g., medical databases, clinical calculators, electronic health records, or simulators). 
The environment returns a response $o_t$:
\begin{equation}
   o_t = \text{Env}(a_t, \mathcal{E}),
\end{equation}
which is fed back into the reasoning loop. 
The trajectory of actions and observations $\tau = \{(a_1, o_1), \ldots, (a_T, o_T)\}$ is then synthesized by the agentic model to generate the final answer:
\begin{equation}
   A = \mathcal{M}_{Agent}(\mathcal{I}, \pi, \tau).
\end{equation}
Therefore, agentic reasoning not only incorporates external medical knowledge but also performs iterative planning and verification, ensuring that the final answer $A$ is both contextually relevant and clinically reliable.

Recent studies in medical AI have focused on integrating agentic reasoning with clinical workflows. 
For example, Mmedagent \cite{li2024mmedagent}, MedRAX \cite{fallahpour2025medrax}, MedOrch(by University of Maryland) \cite{he2025medorch}, and TxAgent \cite{gao2025txagent} employ planning and tool-use mechanisms to guide LLMs in diagnostic decision-making, while 
MedAgent-Pro \cite{wang2025medagent} proposes a step-by-step, evidence-based agentic reasoning paradigm for medical diagnosis.
EHRAgent \cite{shi2024ehragent} empowers LLMs with code generation capabilities to solve complex tabular reasoning tasks over electronic health records.
Similarly, SMR-agents \cite{wang2026smr} and AURA \cite{fathi2025aura} specifically integrate multi-modal medical images for LLM agents to provide decision support, while HiRMed \cite{yang2025tree} introduces RAG into the agent framework for intelligent medical test recommendation.
What's more, Learning to Be A Doctor \cite{zhuang2025learning} views medical LLM agents as graph-structured systems with heterogeneous functional nodes, capable of iterative refinement through diagnostic feedback. 
Furthermore, recent advances emphasize conversational and sequential diagnosis; Tu et al. \cite{tu2025towards} and Nori et al. \cite{nori2025sequential} demonstrate how LLMs can perform iterative information gathering for diagnosis, while Qiu et al. \cite{qiu2025evolving} propose evolving diagnostic agents within virtual clinical environments to refine decision-making strategies.

Beyond single-agent reasoning, multi-agent systems have also been proposed to handle the inherent complexity and specialization of medical tasks. 
For instance, ColaCare \cite{wang2025colacare}, MedAide \cite{wei2024medaide}, MedOrch(by Fudan University) \cite{chen2025mediator}, DoctorAgent-RL \cite{feng2025doctoragent}, MedAgents \cite{tang2023medagents}, MMedAgent-RL \cite{xia2025mmedagent}, MAGDA \cite{bani2024magda}, and MultiMedRes \cite{gu2025proactive} are focused on the framework for multi-agent collaboration, enabling LLM agents to collaborate with each other to solve complex medical tasks.
DruGagent \cite{inoue2025drugagent}, MEDCO \cite{wei2024medco}, and ClinicalAgent \cite{yue2024clinicalagent} utilize multi-agent systems to predict drug-target interaction, medical education, and clinical trial, respectively. Similarly, MAM \cite{zhou2025mam} introduce role-specialized collaboration into the multi-agent framework to solve multi-modal medical diagnosis tasks.
Besides, SeM-Agents \cite{chen2025self} proposes a self-evolving framework for multi-agent medical consultation, while Tree-of-Reasoning \cite{peng2025tree} adopts evidence tree to guide multi-agent reasoning to solve complex medical diagnosis.

\section{Evaluation of Reasoning Capabilities}


The evaluation of medical reasoning systems presents unique challenges that extend beyond traditional NLP metrics to encompass clinical validity, safety, and practical utility. To address this, we structure our evaluation framework in strict alignment with the proposed five-level competency hierarchy, ensuring that assessment protocols are tailored to the distinct cognitive demands of each stratum. As illustrated in Table~\ref{tab:tasks}, our quantitative analysis adopts a stratified approach: foundational capabilities (Levels 1-2) are assessed via precision-based metrics for entity normalization and classification accuracy; diagnostic reasoning (Level 3) is evaluated on logical consistency and differential diagnosis coverage; while the advanced clinical utility of Levels 4-5 necessitates complex, outcome-oriented metrics focusing on decision support quality, safety in treatment planning, and coherence in multi-turn dialogues.

To ensure a rigorous and representative assessment, we curated a diverse set of 18 Large Language Models (LLMs) based on three primary selection criteria: domain specialization, parameter scale, and architectural paradigm.

\begin{itemize}
    \item \textbf{Domain Specialization:} We included both state-of-the-art \textit{General-purpose LLMs} (e.g., GPT-4o, DeepSeek-v3) and specialized \textit{Medical/Clinical LLMs} (e.g., BioGPT, HuatuoGPT). This dichotomy allows us to investigate whether general reasoning capabilities can outperform domain-specific pre-training in complex clinical tasks.
    \item \textbf{Model Scale Diversity:} The selected models span a wide spectrum of parameter sizes, ranging from lightweight models (e.g., BioGPT-347M, Med-Gemma-4B) to massive-scale models (e.g., Kimi-k2-1T). This variation enables an analysis of scaling laws within the medical domain and helps identify the trade-offs between computational efficiency and diagnostic performance.
    \item \textbf{Architectural and Reasoning Paradigms:} Beyond standard dense transformers, our selection incorporates diverse architectures, including Mixture-of-Experts (MoE) and, crucially, the latest reasoning-enhanced models (e.g., o1, DeepSeek-R1-distilled, HuatuoGPT-o1). Including these reasoning models is essential for evaluating performance in high-level tasks (Levels 3-5) that require complex logical chains and multi-step deduction.
\end{itemize}

\begin{table*}[!htbp]
\centering
\resizebox{\textwidth}{!}{
\begin{tabular}{lcccccccccccccccc}
\toprule
\multirow{2}{*}{\bf Models} & \multirow{2}{*}{\bf \#Parameters}& \multicolumn{3}{c}{\bf Level 1} & \multicolumn{3}{c}{\bf Level 2}& \multicolumn{2}{c}{\bf Level 3} & \multicolumn{4}{c}{\bf Level 4}  & \multicolumn{3}{c}{\bf Level 5} \\

\cmidrule(lr){3-5} \cmidrule(lr){6-8} \cmidrule(lr){9-10} \cmidrule(lr){11-14} \cmidrule(lr){15-17}
&
&\makecell{Medical\\ Term Exp.} &\makecell{Medical\\ Term Norm.} & \makecell{\bf Overall} 
&\makecell{Diagnosis\\ Pred.} &\makecell{Urgency\\ Detection} & \makecell{\bf Overall} 
&\makecell{Diagnostic\\ Reasoning} & \makecell{\bf Overall} 
& \makecell{QA\&Decision\\ Support}  & \makecell{Treatment\\ Outcome Pred.} & \makecell{Length of Stay\\ Pred.} & \makecell{\bf Overall} 
& \makecell{Multi-round\\ Dialog} & \makecell{Discharge \\ Summary} & \makecell{\bf Overall} 
\\ \midrule
\multicolumn{17}{c}{\it General-purpose LLMs}\\
\midrule
GPT-oss & 20B &0.222&0.372& 0.297 &0.233&0.329& 0.261 &0.634& 0.634 &0.700&0.644&0.343& 0.597 &0.254&0.425& 0.330 \\
Qwen3 & 235B&0.260&0.350& 0.305 &0.200&\underline{0.642}& 0.332 &0.627& 0.627 &0.694&0.616&\underline{0.377}& 0.595 &0.319&0.412& 0.360 \\
DeepSeek-v3 & 671B &0.242&\underline{0.380}& \underline{0.311} &0.195&0.584& 0.312 &0.572& 0.572 &\underline{0.712}&0.676&\bf 0.411& \bf 0.628 &0.235&0.362& 0.291 \\
Kimi-k2 & 1T &\bf 0.266&0.312& 0.289 &0.272& 0.605& \bf 0.372 &0.637& 0.637 &0.702&\underline{0.720}&0.357& \underline{0.620} &0.235&0.364& 0.292 \\
Mistral & 7B &0.184&0.272& 0.228 &0.172&0.450& 0.255 &0.471& 0.471 &0.360&0.384&0.000& 0.276 &0.116&0.149& 0.131 \\
\midrule
o1 & N/A &\underline{0.258}&0.370& \bf 0.314 &0.210&0.310& 0.240 &\bf 0.655& \bf 0.655 &\bf 0.744&0.668&0.156& 0.578 &\underline{0.398}&0.360& \underline{0.381} \\
DeepSeek-R1-distilled & 8B &0.196&0.316& 0.256 &0.166&0.253& 0.192 &0.519& 0.519 &0.330&0.488&0.140& 0.322 &0.289&0.416& 0.345 \\
QWQ-32B & 32B &0.028&0.050& 0.039 &0.147&0.358& 0.210 &0.529& 0.529 &0.372&0.560&0.100& 0.351 &\bf 0.459&\bf 0.584& \bf 0.514 \\
\midrule
\textbf{\textit{Overall}} & - &0.207 &0.303 & 0.255 &0.199 &0.441 & 0.272 &0.581 & 0.581 &0.577 &0.595 &0.236 & 0.496 &0.288 &0.384 & 0.331 \\
\midrule
\multicolumn{17}{c}{\it Medical/Clinical LLMs}\\
\midrule
BioGPT & 347M &0.020&0.014& 0.017 &0.004&0.000& 0.003 &0.056& 0.056 &0.178&0.032&0.016& 0.101 &0.134&0.416& 0.259 \\
HuatuoGPT & 7B &0.046&0.200& 0.123 &0.055&0.358& 0.146 &0.314& 0.314 &0.200&0.236&0.036& 0.168 &0.054&0.204& 0.120 \\
Med-Gemma & 4B &0.156&\bf 0.408& 0.282 &0.184&\bf 0.713& 0.342 &0.494& 0.494 &0.520&0.476&0.048& 0.391 &0.102&0.188& 0.140 \\
MedicalGPT & 8B &0.138&0.292& 0.215 &0.179&0.508& 0.277 &0.459& 0.459 &0.388&0.252&0.088& 0.279 &0.095&0.190& 0.137 \\
Baichuan-m1 & 14B &0.224&0.372& 0.298 &0.290&0.266& 0.271 &0.492& 0.492 &0.636&0.688&0.096&  0.514 &0.249&0.369& 0.302 \\
\midrule
HuatuoGPT-o1 & 8B &0.152&0.300& 0.226 &0.146&0.292& 0.190 &0.540& 0.540 &0.396&0.352&0.008& 0.288 &0.226&\underline{0.498}&0.346 \\
QoQ-Med & 7B &0.222&0.344& 0.283 &\bf 0.605&0.088& 0.191 &0.602& 0.602 &0.530&0.660&0.096& 0.454 &0.213&0.321& 0.261 \\
ClinicalGPT-r1 & 7B & 0.192&0.260& 0.226 &\underline{0.575}&0.277& 0.337 &\underline{0.640}& \underline{0.640} &0.312&\bf 0.800&0.104& 0.382 &0.165&0.201& 0.181 \\
OpenMedical-R1 & 12B &0.214&0.352& 0.283 &0.345&0.369& \underline{0.364} &0.476& 0.476 &0.558&0.440&0.093& 0.452 &0.271&0.330& 0.297
\\
\midrule
\textbf{\textit{Overall}} & - &0.152 &0.282 & 0.217 &0.265 &0.319 & 0.236 &0.453 & 0.453 &0.413 &0.437 &0.065 & 0.337 &0.168 &0.302 & 0.227 \\
\bottomrule
\end{tabular}
}
\caption{Performance of 18 medical LLMs across five levels of tasks. The best results in each group are \textbf{bolded}, and the second-best results are \underline{underlined}. We performed independent t-tests between general and medical LLMs on the overall scores for each level. The resulting p-values are: Level 1 ($p=0.412$), Level 2 ($p=0.445$), Level 3 ($p=0.072$), Level 4 ($p=0.038$), and Level 5 ($p=0.041$).}
\label{tab:tasks}
\end{table*}

\subsection{General Evaluation Framework}

Evaluating the reasoning capabilities of Large Language Models (LLMs) in medicine requires systematic approaches that capture both output accuracy and reasoning quality. The evaluation process typically encompasses multiple key components including task formulation, model inference, response analysis, performance aggregation, and clinical validation \cite{qiu2025quantifying}. Task formulation structures clinical scenarios into standardized formats, incorporating patient demographics, medical history, symptoms, and test results. During model inference, systems are prompted to provide explicit reasoning chains alongside final outputs, ensuring transparency in decision-making processes. Response analysis examines multiple dimensions including factual accuracy, reasoning completeness, logical consistency, and appropriate handling of uncertainty \cite{reddy2023evaluating}.

The evolution of medical benchmarks has progressed from simple multiple-choice questions to more complex reasoning-focused evaluations. The MMLU-Pro benchmark represents a significant advancement, expanding answer options from four to ten choices while eliminating trivial questions, thereby reducing random guessing likelihood and increasing discriminative power \cite{wang2024mmlu}. 
Similarly, MedHELM \cite{bedi2025medhelm} introduces a holistic evaluation framework covering diverse medical tasks to assess the general utility of LLMs in healthcare.
Studies utilizing these advanced benchmarks have demonstrated that reasoning models achieve impressive accuracy rates, with DeepSeek-R1 reaching 95.1\% on complex medical scenarios after expert reconciliation \cite{guo2025deepseek}. These models exhibit transparency through chain-of-thought reasoning, allowing evaluators to examine logical steps underlying each decision. However, challenges persist in ensuring comprehensive evaluation beyond simple accuracy metrics, as medical decision-making requires consideration of timing, context, and real-world clinical applicability \cite{griot2025large,jin2021disease}.
Specifically, recent frameworks by Hager et al. \cite{hager2024evaluation} and Johri et al. \cite{johri2025evaluation} emphasize the need for evaluating safety in clinical decision-making and the quality of patient interaction tasks beyond pure diagnostic accuracy.


\begin{figure*}
\centering
\includegraphics[width=0.93\linewidth]{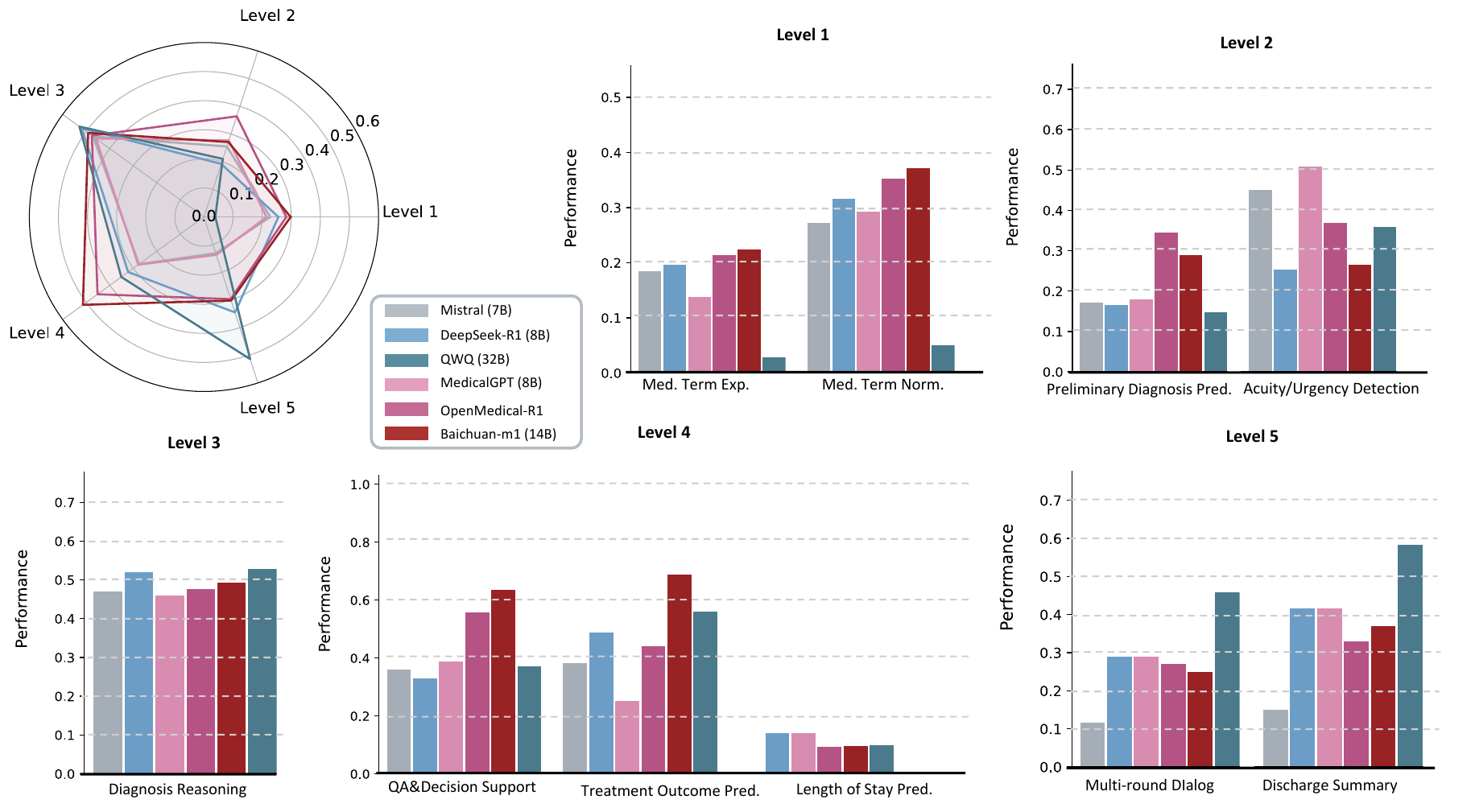}
    \caption{The performance comparison of six SOTA methods on our proposed five-level medical reasoning dataset.}
\end{figure*}

\subsection{Fundamental Evaluation Principles}

\subsubsection{Clinical Validity and Accuracy}

The evaluation of medical reasoning systems must prioritize clinical validity and factual accuracy above all other considerations. Unlike many other AI applications where errors may be inconvenient or embarrassing, errors in medical reasoning can have direct consequences for patient safety and clinical outcomes.
Factual consistency evaluation \cite{tang2023aligning} involves verifying that system outputs align with established medical knowledge, clinical guidelines, and evidence-based practices. This evaluation requires comprehensive medical knowledge bases and expert validation processes to ensure that recommendations are medically sound.
Clinical accuracy assessment \cite{hendrycks2020measuring} examines whether system conclusions and recommendations align with expert clinical judgment across diverse patient scenarios. This evaluation typically involves comparison with expert clinician assessments and analysis of agreement rates across different clinical contexts.
Diagnostic accuracy evaluation \cite{waqas2025reasoning} focuses specifically on the correctness of diagnostic conclusions, including assessment of sensitivity, specificity, positive predictive value, and negative predictive value across different disease categories and patient populations.
Treatment recommendation accuracy \cite{baig2024accuracy} examines whether therapeutic suggestions are appropriate, safe, and evidence-based for specific patient contexts. This evaluation must consider not only the appropriateness of individual recommendations but also their integration into comprehensive treatment plans.

\subsubsection{Reasoning Process Quality}

Beyond accuracy of final outputs, the evaluation of medical reasoning systems must assess the quality and completeness of the reasoning processes that lead to clinical conclusions. This evaluation is crucial for building trust, enabling error detection, and supporting clinical education and decision-making.
Reasoning completeness assessment \cite{hirosawa2023diagnostic} examines whether systems provide comprehensive reasoning chains that include all relevant clinical considerations. Incomplete reasoning can lead to missed diagnoses, inappropriate treatments, or failure to consider important risk factors.
Logical consistency evaluation \cite{loi2024self} verifies that reasoning chains are internally consistent and free from contradictions. Medical reasoning often involves complex relationships between multiple clinical factors, and inconsistencies can indicate fundamental flaws in reasoning processes.
Evidence integration assessment \cite{shool2025systematic} examines how effectively systems incorporate and synthesize multiple sources of clinical evidence, including patient data, medical literature, and clinical guidelines. Effective evidence integration is crucial for comprehensive clinical reasoning.
Uncertainty handling evaluation \cite{savage2024large} assesses how well systems recognize and communicate uncertainty in clinical situations. Medical practice involves significant uncertainty, and systems must be able to acknowledge limitations in their knowledge or confidence.

\subsection{Evaluation Methodologies}

\subsubsection{Human Expert Evaluation}

Human expert evaluation remains the gold standard for assessing medical reasoning systems, providing clinical validation that cannot be achieved through automated metrics alone \cite{ayers2023comparing,johnson2023assessing}. However, expert evaluation is expensive, time-consuming, and subject to inter-rater variability, requiring careful design and implementation.
Clinical expert panels \cite{watson2024eu} provide comprehensive evaluation of system performance across diverse clinical scenarios. These panels typically include clinicians from relevant specialties who evaluate system outputs for accuracy, appropriateness, and clinical utility.
Structured evaluation protocols \cite{seo2024evaluation} ensure consistent and comprehensive assessment across different evaluators and clinical scenarios. These protocols typically include standardized evaluation criteria, rating scales, and guidelines for handling disagreements between evaluators.
Inter-rater reliability assessment \cite{chow2016inter} examines the consistency of evaluations across different expert evaluators. High inter-rater reliability indicates that evaluation results are robust and not dependent on individual evaluator preferences or biases.
Blind evaluation \cite{buhr2023chatgpt} procedures prevent evaluator bias by concealing the source of recommendations being evaluated. This approach enables fair comparison between different systems and between AI systems and human experts.

\subsubsection{Automated Evaluation Metrics}

While human evaluation provides the most clinically relevant assessment, automated metrics enable large-scale evaluation and continuous monitoring of system performance \cite{lee2024analyzing}. These metrics must be carefully designed to capture clinically relevant aspects of performance.
Accuracy-based metrics, including precision, recall, F1-score, and area under the ROC curve, provide quantitative assessment of system performance on classification and prediction tasks. These metrics are particularly useful for diagnostic reasoning and risk prediction applications.
Clinical validity metrics \cite{alaa2025medical} assess whether system outputs conform to established medical knowledge and guidelines. These metrics can be partially automated through comparison with structured medical knowledge bases and clinical decision rules.
Reasoning chain evaluation \cite{nachane2024few} metrics assess the quality of step-by-step reasoning processes, including measures of logical consistency, completeness, and evidence integration. These metrics require sophisticated natural language processing techniques to analyze reasoning quality.
Safety and harm prevention metrics \cite{gilson2023does,kung2023performance} identify potentially dangerous or inappropriate recommendations that could lead to patient harm. These metrics are crucial for ensuring that systems meet safety requirements for clinical deployment.

\subsection{Five-Level Medical Reasoning Benchmark}

To systematically evaluate LLM medical reasoning capabilities according to our proposed competency hierarchy, we have constructed a benchmark dataset that spans all five levels of medical reasoning competence. This benchmark provides structured assessment aligned with both theoretical frameworks and practical clinical needs.

\subsubsection{Dataset Composition and Statistics}




\paragraph{Data Construction and Preprocessing.}
The benchmark dataset comprises 5,000 carefully selected samples, balanced with 1,000 samples for each competency level. Samples are aggregated from multiple source datasets to ensure diversity and prevent overfitting to specific data distributions. Prior to inclusion, a rigorous data cleaning pipeline was applied. This involved: (1) De-identification, ensuring all patient information was anonymized in compliance with HIPAA/GDPR standards; (2) Format Standardization, unifying heterogeneous inputs (e.g., SOAP notes, discharge summaries) into a consistent structured format; and (3) Quality Filtering, where incomplete or ambiguous cases were removed via rule-based heuristics. The random selection process within each level ensures comprehensive representation across medical specialties (internal medicine, surgery, pediatrics), task formats, and patient demographics.

\paragraph{Annotation Procedures.}
To construct the benchmark inputs, we adopted a model-assisted pipeline to ensure scalability while maintaining clinical relevance. Specifically, we utilized Large Language Models (LLMs) to extract key clinical features and context from existing source benchmarks, restructuring them into a standardized input format. The diagnostic labels or clinical answers were directly inherited from the original ground truth of the source datasets to preserve the expert-validated accuracy. This process ensures that while the input presentation is standardized and enriched by the LLM, the core medical correctness relies on established gold-standard data.

\paragraph{Inter-Annotator Agreement and Quality Control.}
Given the semi-automated nature of the dataset construction, human effort was primarily directed toward quality assurance and validation rather than de novo annotation. To quantify the reliability of the LLM-generated inputs and their alignment with the ground truth labels, we implemented a rigorous review protocol. A subset of 20\% of the data was subjected to a double-blind review, where two independent clinicians assessed the samples for factual consistency, input clarity, and label alignment without access to each other's evaluations. The Inter-Annotator Agreement (IAA) on the acceptance of these samples was calculated using Cohen's Kappa, yielding a score of $\kappa = 0.88$. This high degree of concordance confirms the robustness of our pipeline, verifying that the automated reconstruction preserved clinical integrity without introducing hallucinations or semantic distortions.

\subsubsection{Experimental Analysis on Reasoning Capabilities}
We analyze performance along our five-level competency hierarchy, contrasting general-purpose LLMs with medical/clinical LLMs and highlighting where specific models excel.

\paragraph{Level 1: Medical term understanding}
For Term Expansion, the best model is \texttt{Kimi-k2}, narrowly ahead of \texttt{o1}; among medical LLMs, \texttt{Baichuan-m1} leads its cohort but still trails the top general models. 
For Term Normalization, the medical \texttt{Med-Gemma} is strongest, edging \texttt{DeepSeek-v3}. 
Overall, general LLMs average higher on Level 1 expansion (\textit{Overall}: 0.207 vs. 0.152), reflecting broad lexical coverage; however, targeted medical pretraining can surpass general models on normalization (domain-specific ontology alignment).

\paragraph{Level 2: Clinical signal recognition}
Medical-specialist models dominate Diagnosis Prediction. The \texttt{QoQ-Med} and \texttt{ClinicalGPT-R1} clearly outperform all general models.
Conversely, Urgency Detection is led by the lighter medical \texttt{Med-Gemma} but with \texttt{Qwen3} close behind.
The General cohort averages higher on this task, and these results suggest domain adaptation is crucial for mapping symptoms to likely diagnoses, whereas urgency cues often benefit from broad world and linguistic priors captured by frontier general LLMs.

\paragraph{Level 3: Diagnostic reasoning}
\texttt{ClinicalGPT-R1} attains the top score. Despite this medical win at the head, the cohort average favors general models, implying that strong stepwise reasoning remains more consistent in general LLMs, while select medical models can peak when their pretraining aligns with diagnostic COT.

\paragraph{Level 4: Decision support and structured predictions}
For QA \& Decision Support, \texttt{o1} is best, with \texttt{DeepSeek-v3 } second. For Treatment Outcome Prediction, \texttt{ClinicalGPT-r1} leads decisively, with \texttt{Kimi-k2} next. For Length-of-Stay (LOS) prediction, \texttt{DeepSeek-v3} is best. 
General models strongly outperform medical models on average in all three Level~4 tasks. Notably, LOS is challenging for all systems (low absolute numbers), reflecting difficulties in learning calibrated, structured temporal outcomes from text alone.

\paragraph{Level 5: Clinical signal recognition}
\texttt{QWQ-32B} excels at Multi-round Dialog andDischarge Summary, with \texttt{o1} and \texttt{HuatuoGPT-o1} as runners-up. Despite some strong medical entrants (e.g., HuatuoGPT-o1 for summarization), general models hold the advantage on average, likely due to their richer conversational dynamics and broader stylistic control. 
Despite weak scores elsewhere, an interesting outlier is \texttt{BioGPT}, which shows comparatively solid discharge summarization, indicating the value of narrow biomedical language modeling for document-style generation.

Taken together, these results suggest a practical division of labor. When tasks require precise clinical priors and particularly diagnosis-centric prediction. 
Domain-specialized models provide clear gains. When tasks demand broad reasoning, instruction adherence, or extended discourse, high-capacity general LLMs remain stronger on average. 
Parameter scale alone does not explain the frontier. Instruction tuning quality and domain-constrained curricula are at least as important, and structured clinical predictions (e.g., LOS) expose current limits in temporal reasoning and calibration. 
A system-level implication is to adopt a routing or mixture-of-experts strategy that dispatches diagnosis-heavy queries to medical models (e.g., QoQ-Med, ClinicalGPT-r1) while leveraging general models (e.g., o1, DeepSeek-v3, Kimi-k2, Qwen3, QWQ-32B) for decision support, dialogue, and summarization.

\begin{table*}[t]
\centering
\small
\resizebox{\textwidth}{!}{
\begin{tabular}{cccc}
\toprule
Level Task & Model & Model Output & Failure Mode \\
\midrule

\multirow{4}{*}{\makecell{\textbf{Level 1} \\ Term \\ Normalization}}
& \texttt{o1} 
& \makecell[c]{\emph{``Standardized term: Postmenopausal Bleeding''} \\ (Ground truth: \emph{``Vaginal Haemorrhage''}) } 
& Ontology Misalignmen \\
\cmidrule{2-4}
& \texttt{BioGPT} 
& \makecell[c]{\emph{``Standardized term: Chronic Abdominal Pain''} \\ (Ground truth: \emph{``Back Pain''}) } 
& Ontology Misalignmen \\

\midrule

\multirow{4}{*}{\makecell{\textbf{Level 2} \\ Diagnosis \\ Prediction}}
& o1 
& \makecell[c]{\emph{``Final diagnosis: Pneumonia''} \\ (Case labeled as \emph{Healthy})} 
& Brittle Diagnosis \\
\cmidrule{2-4}
& BioGPT 
& \makecell[c]{\emph{``Final diagnosis: Acute bronchitis''} \\ (Case labeled as \emph{Pulmonary Embolism})} 
& Brittle Diagnosis \\

\midrule

\multirow{4}{*}{\makecell{\textbf{Level 3} \\ Diagnostic \\ Reasoning}}
& o1
& \makecell[l]{
    \emph{``Positive''} (Cardiovascular disease diagnosis  despite normal blood pressure,  
    normal \\ cholesterol and glucose, no smoking or alcohol use, and an active lifestyle)
  } 
& Brittle Diagnosis \\
\cmidrule{2-4}
& BioGPT 
& \makecell[l]{
    \emph{``Negative''} (Cardiovascular disease diagnosis despite positive ground truth 
    with  \\ age-related risk and active smoking history)
} 
& Brittle Diagnosis \\

\midrule

\multirow{4}{*}{\makecell{\textbf{Level 4} \\ LOS \\ Prediction}}
& \texttt{o1} 
& \makecell[c]{ 
\emph{``Predicted LOS: 4 days''} \\
(valid JSON format and fluent rationale, but potential temporal under-specification) 
} 
& Temporal Under-Specification \\
\cmidrule{2-4}
& BioGPT 
& \makecell[c]{
\emph{``If the patient was male, he was female, and if the patient was on dialysis, he was male.''} \\
(logical inconsistency and loss of patient attribute coherence) 
} 
& Logical Inconsistency \\

\midrule

\multirow{4}{*}{\makecell{\textbf{Level 5} \\ Discharge \\ Summary}}
& \texttt{o1} 
& \makecell[c]{ 
\emph{``Upon discharge, the patient’s condition was stable \ldots''} \\
(coherent narrative flow with risk of inferred or reordered events) 
}  
& Narrative Over-Interpretation \\
\cmidrule{2-4}
& BioGPT 
& \makecell[c]{ 
\emph{``Patient Condition: \ldots\ History of Present Illness: \ldots\ Discharge Instructions: \ldots''} \\
(template-style listing with limited narrative integration)
} 
& Rigid Templating \\

\bottomrule
\end{tabular}}
\caption{Representative failure cases across Levels~1--5.
Failures at Levels~1--3 are localized, including ontology misalignment in term normalization and brittle diagnostic reasoning under non-canonical symptom patterns.
In contrast, Levels~4 and~5 expose more fundamental limitations related to execution fidelity, temporal grounding, and discourse-level control.}
\label{tab:failure_cases_l1_l5}
\end{table*}

\paragraph{Failure case analysis and qualitative findings}
Failure case analysis reveals systematic and level-dependent limitations that are not fully captured by aggregate metrics. 
At Levels~1--3, observed failures are comparatively localized: general LLMs occasionally misalign surface medical terms with formal ontologies during normalization, while medical LLMs exhibit brittle diagnostic reasoning when symptom patterns deviate from canonical clinical templates. These errors, however, remain bounded and do not dominate overall performance trends.

In contrast, Level~4 structured prediction tasks, particularly Length-of-Stay (LOS), expose pronounced divergences between model families. 
Medical LLMs frequently fail at the execution level. For example,  BioGPT occasionally generates logically inconsistent text. These behaviors indicate difficulties in aligning clinical safety constraints with strict output specifications.
By contrast, general-purpose LLMs such as \texttt{o1} reliably adhere to required JSON formats and generate fluent reasoning traces, even when their LOS predictions are inaccurate. This execution robustness largely explains the statistically significant performance advantage of general models at Level~4: they function as competent task executors, whereas medical LLMs often fail before the correctness of the prediction can be meaningfully evaluated.

Level~5 discharge summarization further highlights qualitative differences in generative behavior. 
General LLMs produce summaries with strong narrative flow, organizing clinical events into coherent, temporally ordered stories (e.g., ``Upon discharge, the patient\ldots''), closely resembling real-world clinical documentation. In contrast, medical LLMs such as BioGPT favor rigid, list-like structures (e.g., ``Patient Condition:'', ``History:'') that preserve factual elements but lack integrative writing and discourse-level synthesis.
While this narrative capacity contributes to higher average scores for general models, it also introduces risks of over-interpretation, including event reordering or inferred causal relations not explicitly supported by the source text.

Taken together, these failure cases suggest that higher-level clinical tasks are constrained not only by medical knowledge coverage, but by execution fidelity, format adherence, and discourse-level control. 
The results reinforce a system-level design principle: domain-specialized medical models are best suited for diagnosis-centric and safety-critical subtasks, while general LLMs are more reliable for structured execution, dialogue management, and narrative clinical documentation.

\begin{figure*}
\centering
    \includegraphics[width=0.93\linewidth]{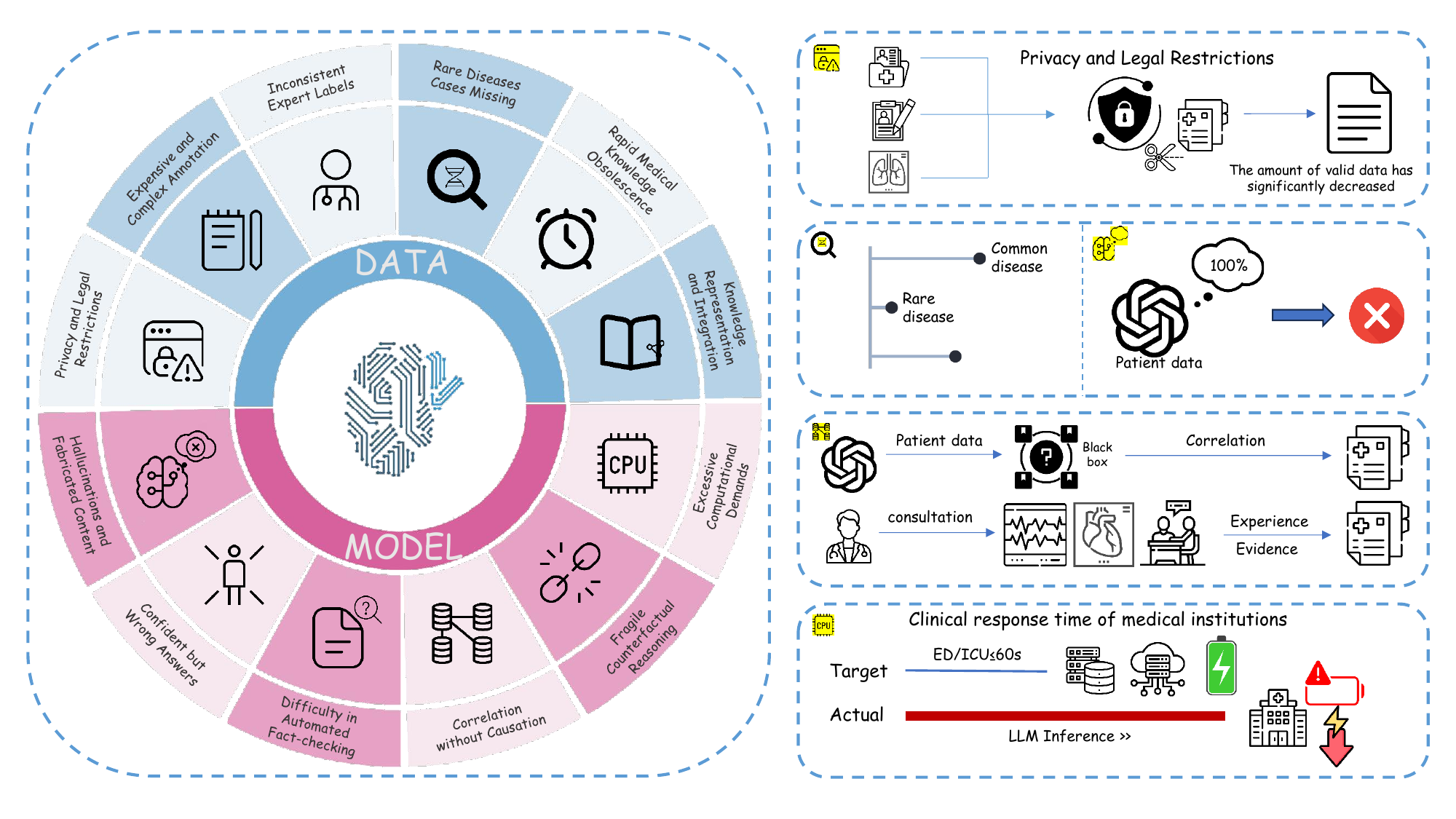}
    \caption{Data and knowledge challenges in medical reasoning with large language models, and a comparison between correlation-based model diagnosis and causal, evidence-grounded clinical reasoning in chest pain cases.}
    \label{fig:challenge}
\end{figure*}

\section{Challenges and Limitations}
Despite significant advances in medical reasoning systems, numerous challenges and limitations continue to constrain their development, deployment, and clinical impact \cite{yu2018artificial, al2023review, secinaro2021role}. 
Understanding these challenges is crucial for directing future research efforts and setting realistic expectations for clinical implementation.

\subsection{Data and Knowledge Challenges}

\subsubsection{Data Scarcity and Quality Issues}
The effective development of medical reasoning systems is fundamentally constrained by the availability of large-scale, high-quality, and annotated training data. As shown in Fig.~\ref{fig:challenge}, medical data faces severe challenges in terms of annotation complexity, privacy protection, reliance on domain expertise, timeliness, and data imbalance \cite{li2021critical}.
High-quality medical data not only requires professionals to spend substantial time and effort on annotation, but also suffers from issues such as privacy and regulatory restrictions, severe shortage of rare disease samples, and rapid medical knowledge obsolescence. More seriously, differences in judgment among experts often make it difficult to form a unified “gold standard” \cite{murdoch2021privacy, thapa2021precision}.
In some complex reasoning tasks, such as diagnostic reasoning or treatment planning, annotation requires not only the correct answers but also detailed reasoning chains to reflect the expert’s thought process, which is costly and time-consuming. In addition, certain privacy and regulatory restrictions (such as the HIPAA regulatory process in the United States, shown in Fig.~\ref{fig:challenge}) further reduce the size and diversity of datasets. What's more, data imbalance is particularly prominent in medical applications; as shown in Fig.~\ref{fig:challenge}, rare diseases and uncommon cases are severely underrepresented in training data, leading to poor performance of medical reasoning systems in these clinically important scenarios. In addition, with the rapid evolution of medical knowledge and best practices, outdated medical data may cause reasoning systems to recommend obsolete therapies or fail to incorporate the latest reasoning guidelines. Furthermore, in complex clinical scenarios, inter-annotator agreement (IAA) can be very low, making it difficult to establish the “gold standard” required for training and evaluation \cite{zajkac2023ground}.
Therefore, the development of medical reasoning systems is severely constrained by data scarcity and quality issues.

\subsubsection{Knowledge Representation and Integration}
Medical reasoning needs to combine different types of knowledge, including medical facts, clinical guidelines, research literature, and clinical experience. However, there remain significant challenges in how to represent and integrate this knowledge \cite{riano2019ten}.
For example, existing medical knowledge bases (such as "UMLS" and "SNOMED CT"), although broad in coverage, are insufficient in capturing the finer semantics and specific contexts required in clinical reasoning \cite{fung2005integrating}.
Moreover, clinical guidelines are usually written in natural language, often containing ambiguity, exceptions, or even conflicting recommendations, which makes transforming them into machine-usable formats both complex and error-prone. In addition, the scale of medical literature is vast and continuously growing, and current systems remain limited in their ability to filter, evaluate, and integrate evidence. Meanwhile, clinical experience often depends on specific contexts and carries implicit characteristics, making it even harder to formalize and utilize. Overall, medical reasoning not only relies on factual knowledge but also requires the integration of guidelines, literature, and experience, and how to effectively represent and combine these diverse sources of knowledge remains a central challenge for medical reasoning systems.

\subsection{Model Limitations and Technical Challenges}

\subsubsection{Hallucination and Factual Accuracy}
One prominent risk of large language models in medical applications is that they may generate content that appears reasonable but is in fact incorrect, a phenomenon known as hallucination \cite{ji2023survey}.
In medical scenarios, such errors may have serious consequences for patient safety and clinical decision-making. This problem is mainly reflected in four aspects: First, the erroneous content of model hallucinations can take many forms. Hallucinations may appear as incorrect drug dosages, fabricated diseases, inappropriate treatment recommendations, or even invented research citations that do not exist \cite{agarwal2024medhalu, kim2025medical}.  
Such content is often highly deceptive and can easily mislead non-professional users. Second, the confidence calibration of the model is insufficient. Medical reasoning systems often perform poorly in judging the reliability of their own answers. As shown in Fig.~\ref{fig:challenge}, they may exhibit excessive confidence in incorrect answers, which leads users to over-rely on them and increases risk. Third, automated fact-checking is extremely difficult. Medical knowledge is not only complex but also highly context-dependent. Automated fact-checking must simultaneously determine whether medical facts are correct, clinically applicable, and valid in specific contexts, which makes the verification process very challenging \cite{sarrouti2021evidence}. 
Finally, the issue of insufficient traceability and lack of evidence support also deserves attention. Large language models often generate content without authoritative sources, making it difficult for clinicians to verify the information and to build sufficient trust in the model’s output. In summary, the problems of hallucination and factual accuracy constitute one of the core challenges of medical reasoning models and are key obstacles limiting their safe clinical application.

\subsubsection{Reasoning Depth and Complexity}
Existing medical reasoning systems often remain at the level of shallow pattern matching and lack a deep understanding of medical principles and causal relationships \cite{holzinger2019causability}. 
This deficiency is mainly reflected in three aspects: First, the causal reasoning ability of diagnostic models is limited. Models often only identify correlations in the data but fail to grasp the underlying causal mechanisms, which may lead to incorrect recommendations based on spurious correlations. Second, multi-step reasoning chains are unstable. In complex tasks that require multi-step logical chains, systems often lose coherence or exhibit logical breaks when integrating information from multiple sources. In addition, uncertainty handling and counterfactual reasoning are also highly insufficient. Since medical reasoning involves multiple uncertainties, systems still face difficulties in quantifying, propagating, and interpreting these uncertainties. At the same time, existing models lack counterfactual reasoning ability (i.e., considering alternative scenarios and their effects), which is particularly critical in treatment planning and risk assessment \cite{yang2023counterfactual, prosperi2020causal}.
A typical case is the differential diagnosis of chest pain, as shown in Fig.~\ref{fig:challenge}. Angina pectoris and gastroesophageal reflux overlap greatly in symptoms (such as retrosternal pressure-like discomfort). Data-driven correlation models may directly map “chest pain and nocturnal or postprandial discomfort” to a single diagnosis, resulting in hallucination-like misjudgment. In contrast, clinicians analyze along the causal chain the inducing and relieving factors (for example, exertion-induced, relieved by rest or nitrates indicating myocardial ischemia; aggravated when lying down or after meals and accompanied by acid reflux indicating gastroesophageal reflux), and then combine evidence such as ECG, troponin, or proton pump inhibitor trial to verify and produce a more accurate diagnosis. This contrast shows that systems lacking pathophysiological causal modeling and evidence traceability are easily misled by superficial correlations and cannot ensure reliability in critical clinical scenarios.

\subsubsection{Scalability and Computational Requirements}
Complex medical reasoning systems often require enormous computational power, which is in natural conflict with the clinical demand for efficient applications \cite{wahl2018artificial}. 
In some routine scenarios, slower inference speed may only reduce the user experience, but as shown in Fig.~\ref{fig:challenge}, in emergency and critical care settings where real-time response is required, excessively long inference times can directly affect patient safety. Furthermore, medical resource limitations make the problem even more prominent. Many hospitals lack sufficient computing infrastructure and budget, while resource-limited regions (such as rural or low-resource hospitals), although in greatest need of AI support, are least able to afford the deployment and maintenance costs of high-computation systems. Beyond this, sustainability issues cannot be ignored. Large-scale model inference not only consumes enormous energy but also brings long-term environmental burdens and operational pressures \cite{strubell2020energy}. 
Therefore, from conflicts with real-time requirements, to resource insufficiency, and further to energy consumption and sustainability, the computational bottleneck has gradually become the most severe obstacle to the clinical implementation of medical reasoning systems.

\section{Recent Advances and Future Directions}

The field of medical reasoning with large language models is rapidly evolving, with significant advances occurring across multiple dimensions including technical capabilities, clinical applications, and regulatory frameworks. This section examines recent breakthroughs and identifies promising directions for future research and development.

\subsection{Advanced Reasoning Architectures}

Recent developments in reasoning architectures have significantly enhanced the capabilities of medical reasoning systems, moving beyond simple question-answering toward more sophisticated clinical reasoning processes.
Long-form chain-of-thought reasoning \cite{chen2025towards}  has emerged as a particularly promising approach for medical applications, enabling systems to engage in extended reasoning processes that more closely mirror expert clinical thinking. Models like OpenAI's o1 and DeepSeek-R1 have demonstrated that extended reasoning can significantly improve performance on complex reasoning tasks, and early applications to medical domains show similar promise. The integration of tool-augmented reasoning \cite{ma2024sciagent} with medical knowledge bases \cite{he2025medorch} and clinical decision support tools has enabled more accurate reasoning. These systems can access up-to-date medical information, perform calculations, and integrate multiple sources of evidence during the reasoning process. Self-reflection and error correction mechanisms \cite{ji2023towards} have been developed to enable systems to review and refine their reasoning processes. These capabilities are particularly important in medical contexts where accuracy is paramount and errors can have serious consequences. Multi-agent reasoning systems \cite{vicari2003multi} that combine multiple specialized models or reasoning approaches have shown promise for complex medical scenarios that require diverse types of expertise. These systems can leverage the strengths of different models while mitigating individual weaknesses.

\subsection{Multimodal Integration Advances}

The integration of multiple data modalities has seen significant advances, enabling more comprehensive and accurate medical reasoning.
Vision-language integration has reached new levels of sophistication, with models like GPT-4V \cite{yang2023dawn}, DeepMedix-R1~\cite{lin2025foundation}, and Med-Gemini \cite{saab2024capabilities} demonstrating impressive capabilities in analyzing medical images and integrating visual information with clinical text. These advances enable reasoning about radiology images, pathology slides, and clinical photographs in conjunction with textual clinical information.
Time-series integration capabilities \cite{li2014physiological} have improved significantly, enabling better reasoning about physiological monitoring data, laboratory trends, and disease progression patterns. These capabilities are particularly important for intensive care and chronic disease management applications.
Structured data integration \cite{wu2025medreason} has advanced to enable seamless reasoning across electronic health record data, laboratory results, and clinical documentation. These advances enable more comprehensive clinical reasoning that considers all available patient information.
Genomic data integration \cite{ioannidis2018evidence} represents an emerging frontier, with early systems demonstrating the ability to integrate genetic information with clinical data for personalized medicine applications.

\subsection{Safety and Reliability Improvements}

Significant advances have been made in improving the safety and reliability of medical reasoning systems, addressing critical concerns about clinical deployment.
Uncertainty quantification methods \cite{atf2025challenge} have become more sophisticated, enabling systems to provide more accurate estimates of their confidence in different recommendations. These improvements help clinicians understand when system outputs should be trusted and when additional verification is needed.
Hallucination detection and mitigation techniques \cite{kim2025medical} have been developed specifically for medical applications, using domain-specific knowledge bases and fact-checking mechanisms to identify and prevent factually incorrect outputs. 
Adversarial robustness \cite{chen2025cares} has been improved through specialized training techniques that make systems more resistant to inputs designed to elicit incorrect or harmful responses.
Safety guardrails and content filtering mechanisms \cite{yang2024ensuring} have been developed to prevent systems from generating potentially harmful recommendations, such as dangerous drug combinations or inappropriate treatment suggestions.

\section{Conclusion}

This survey has addressed the growing challenges in medical decision-making, where clinicians must process large volumes of data while managing uncertainty and complex trade-offs. Through our dual-view framework linking clinical competencies with reasoning types, we have clarified what kinds of reasoning are actually needed in clinical practice and how current LLM capabilities map to these specific medical needs across the care pathway.

Our five-level competency framework reveals that current models perform well on knowledge recognition and classification tasks but face important challenges with higher-level reasoning requiring integration, personalization, and dynamic interaction. The evaluation of 18 state-of-the-art models on our benchmark demonstrates a clear performance pattern: medical specialist models excel in diagnosis-centric tasks while general models show stronger performance in decision support, dialogue, and summarization scenarios. Notably, temporal predictions such as length-of-stay remain difficult for all systems, and model size alone does not explain performance differences. Instruction quality, domain-specific data, and reasoning training appear equally important.

These findings point toward practical deployment strategies that route diagnosis-heavy queries to specialist models while using general models for supportive tasks. However, key limitations remain in areas such as uncertainty handling, grounding to trusted medical sources, and integration into real clinical workflows. Moving forward, the field requires better evaluation frameworks that assess both accuracy and reasoning quality, improved training approaches that combine domain expertise with robust reasoning capabilities, and careful attention to safety and workflow integration.

\section*{Statements and Declarations}

The authors have no relevant financial or non-financial interests to disclose.

\section*{Acknowledgments}

This work was supported by the Guangdong Provincial Fund for Basic and Applied Basic Research—Regional Joint Fund Project (Key Project) (No. 2023B1515120078), the National Natural Science Foundation of China (No. 62476097), the Science and Technology Planning Project of Guangdong Province (No. 2025B0101120003), the Guangdong Provincial Natural Science Foundation for Outstanding Youth Team Project (No. 2024B1515040010), the Fundamental Research Funds for the Central Universities, South China University of Technology (No. x2rjD2250190), the National Natural Science Foundation of China (No. 62372314), the 1.3.5 project for disciplines of excellence from West China Hospital of Sichuan University (No. 2023HXFH044), and the Hong Kong Research Grants Council under the Theme-based Research Scheme (Project No. T22-501/23-R).
The experimental part of this work was supported by The Centre for Large AI Models (CLAIM) of The Hong Kong Polytechnic University.








\bibliography{references}

\clearpage

\begin{figure}[h]%
\centering
\includegraphics[width=0.3\textwidth]{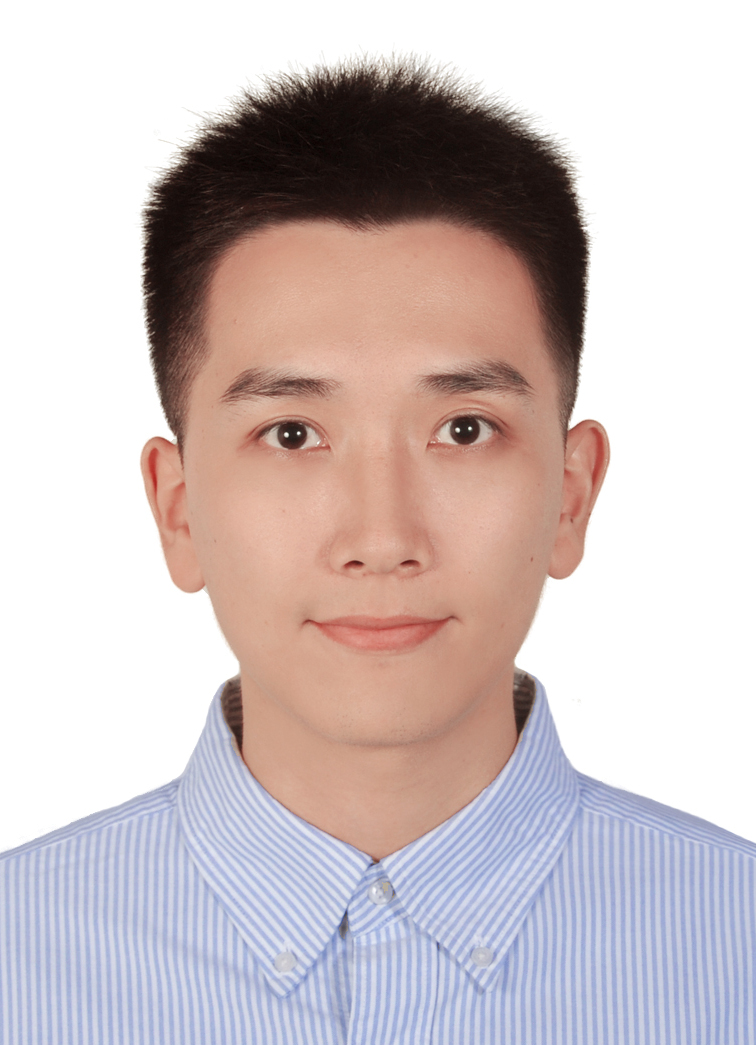}
\end{figure}


\noindent{\bf Qi Peng}\quad is currently a dual Ph.D. student at Hong Kong Polytechnic University and South China University of Technology, supervised by Prof. Qing Li and Prof. Yi Cai. He received both his Master’s degree and Bachelor’s degree in South China University of Technology in 2022 and 2019, respectively. His research interests include large language models, multi-agent systems, and intelligent healthcare. His research works have been published as the first author in ACM MM, IEEE TMI, and KBS.

E-mail: qi0125.peng@connect.polyu.hk 

ORCID iD: 0000-0001-7747-3293

\begin{figure}[h]%
\centering
\includegraphics[width=0.3\textwidth]{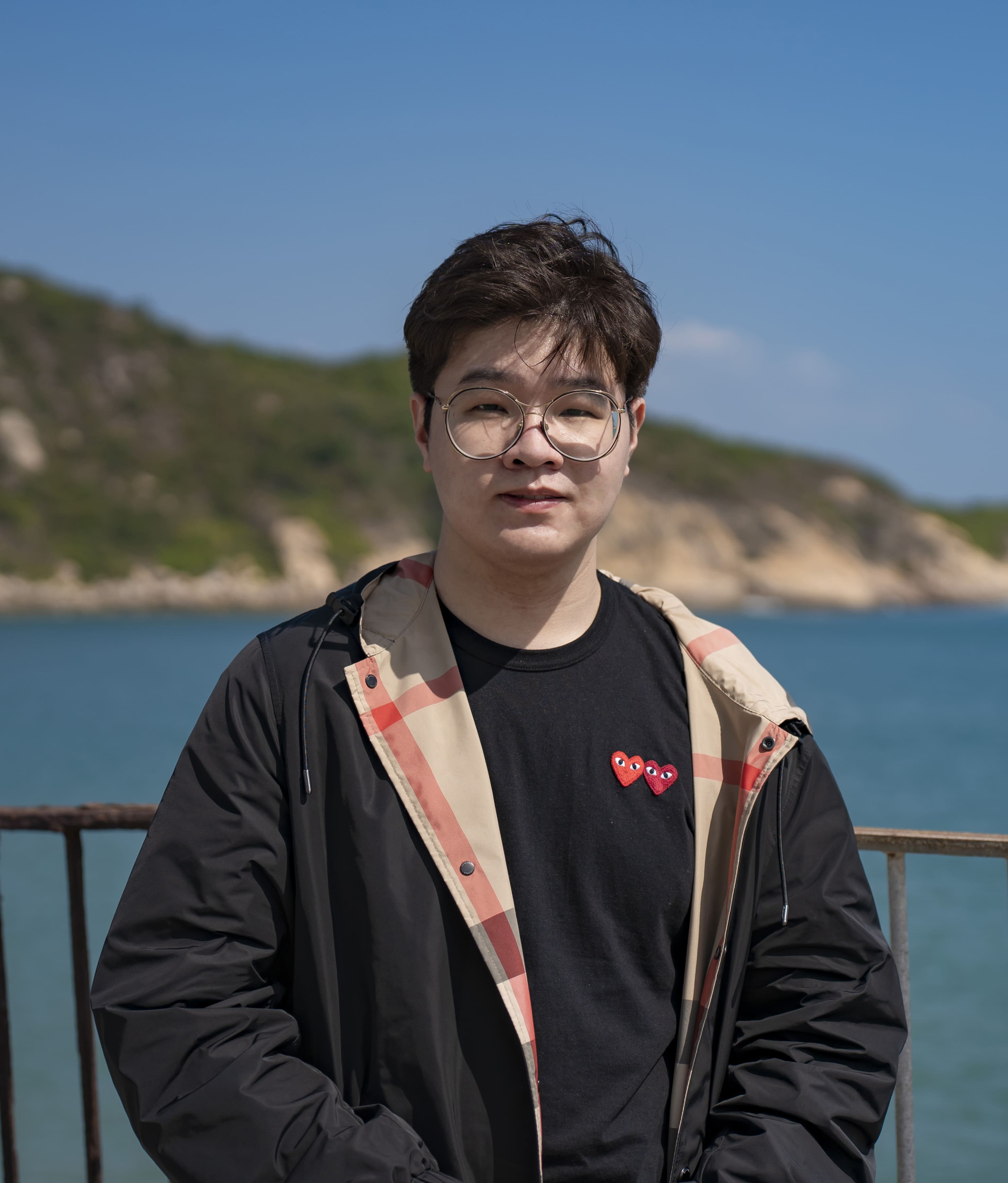}
\end{figure}

\noindent{\bf Jiatong Li}\quad is currently a PhD candidate of the Department of Computing (COMP), The Hong Kong Polytechnic University (funded by HKPFS). Before joining the PolyU, he received his Master's degree of Information Technology (with Distinction) from the University of Melbourne. In 2021, he got his bachelor's degree in Information Security from Shanghai Jiao Tong University. His interest lies in Natural Language Processing, Drug Discovery, Medicine, and healthcare. He has published innovative works in top-tier conferences and journals such as IJCAI, ACL, CVPR, and IEEE TKDE. For more information, please visit https://phenixace.github.io/.

ORCID iD: 0000-0001-7705-2296.

\begin{figure}[h]%
\centering
\includegraphics[width=0.3\textwidth]{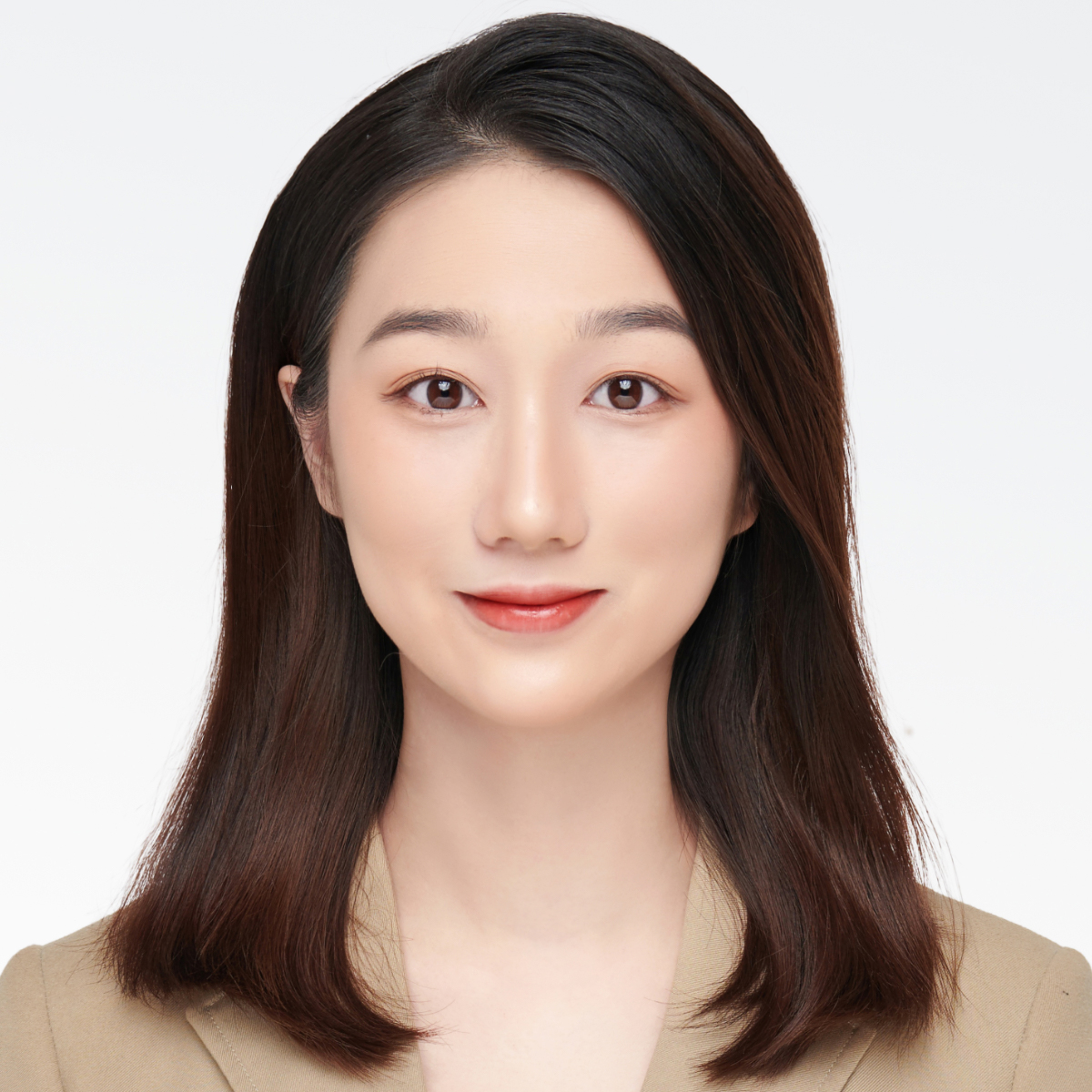}
\end{figure}

\noindent{\bf Sirui Huang}\quad
received the MSc degree from Electronic and Electrical Engineering Department, University College London, London, U.K., in 2021. She is currently a Ph.D. student in the Computing Department at Hong Kong Polytechnic University, Hong Kong, China, supervised by Prof. Qng Li and Prof. Guandong Xu. Her research interests mainly include recommender systems, large language models and table reasoning. 

ORCID iD: 0000-0003-1206-2260.

\begin{figure}[h]%
\centering
\includegraphics[width=0.3\textwidth]{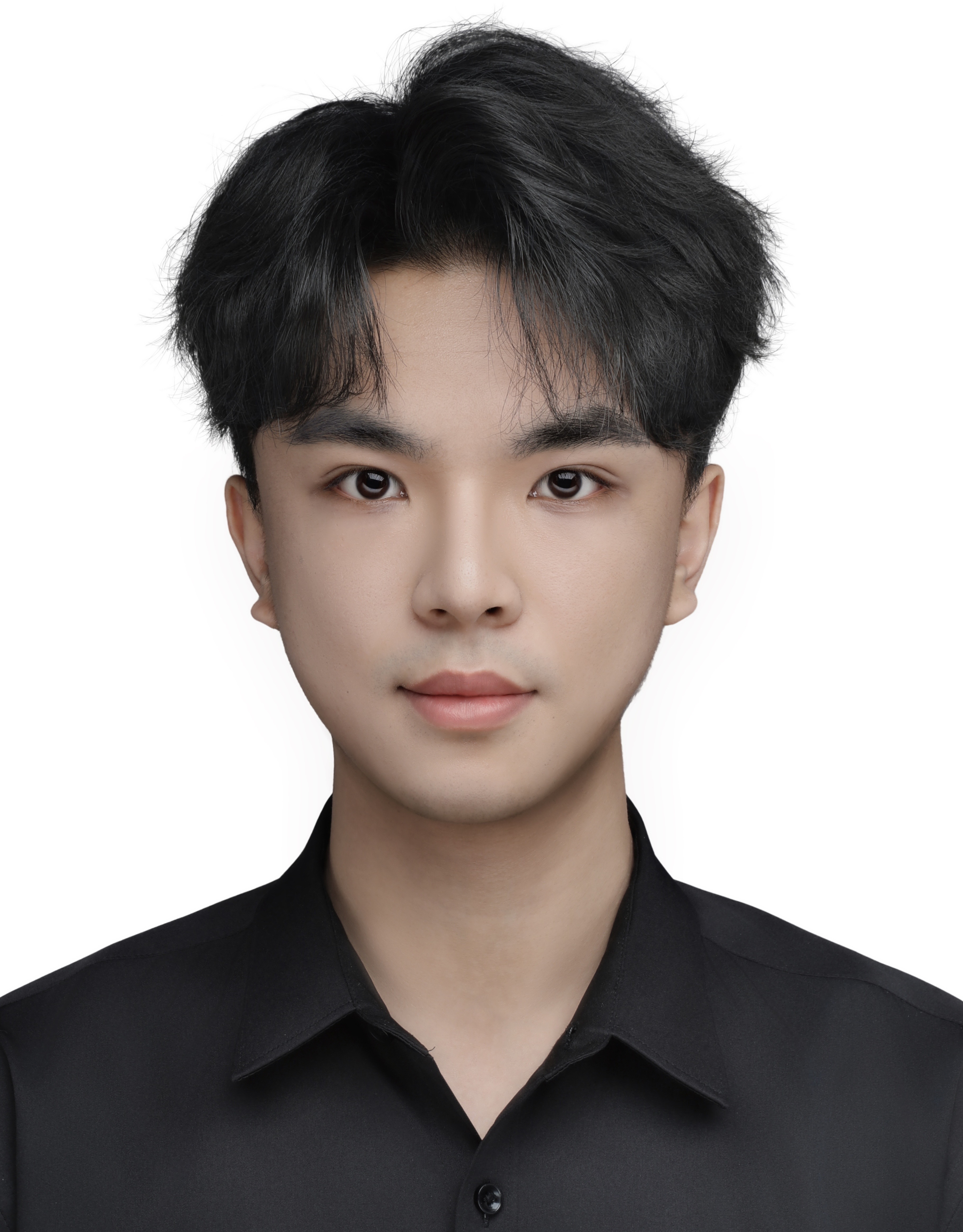}
\end{figure}

\noindent{\bf Yiyang Jiang}\quad received the B.Sc. (Hons., 1st Class) degree in Computer Science from The Hong Kong Polytechnic University in 2024. He is currently a Ph.D. student in the Department of Computing at The Hong Kong Polytechnic University, supported by the Hong Kong PhD Fellowship Scheme from the Research Grants Council. His research interests include vision-language understanding, natural language processing, large language models, and health computing.

ORCID iD: 0009-0007-1169-4465.

\clearpage

\begin{figure}[h]%
\centering
\includegraphics[width=0.3\textwidth]{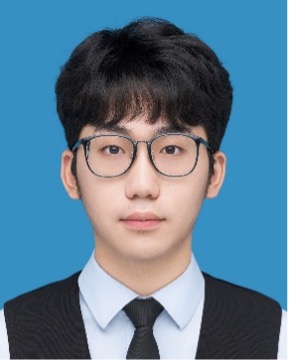}
\end{figure}

\noindent{\bf Kaisong Gong}\quad received the B.S. degree from Tianjin University in 2024. He is currently a graduate student at The University of Hong Kong and a research assistant at The Hong Kong Polytechnic University. His research interests include VLLM reasoning and parameter-efficient fine-tuning. 

E-mail: gks70853@connect.hku.hk,

ORCID iD: 0009-0008-7612-3177.

\begin{figure}[h]%
\centering
\includegraphics[width=0.3\textwidth]{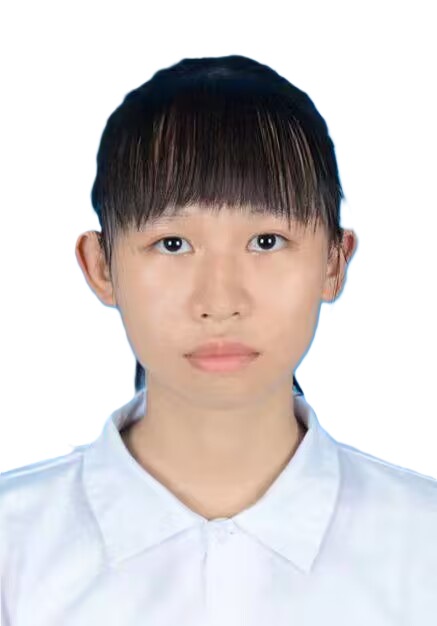}
\end{figure}

\noindent{\bf Ronger Ding}\quad is an undergraduate student at the Software College of South China University of Technology. Her research interest include the application of large language models in various fields.

E-mail: se30480335ding@mail.scut.edu.cn.

ORCID iD: 0009-0005-6422-1130.

\clearpage

\begin{figure}[h]%
\centering
\includegraphics[width=0.3\textwidth]{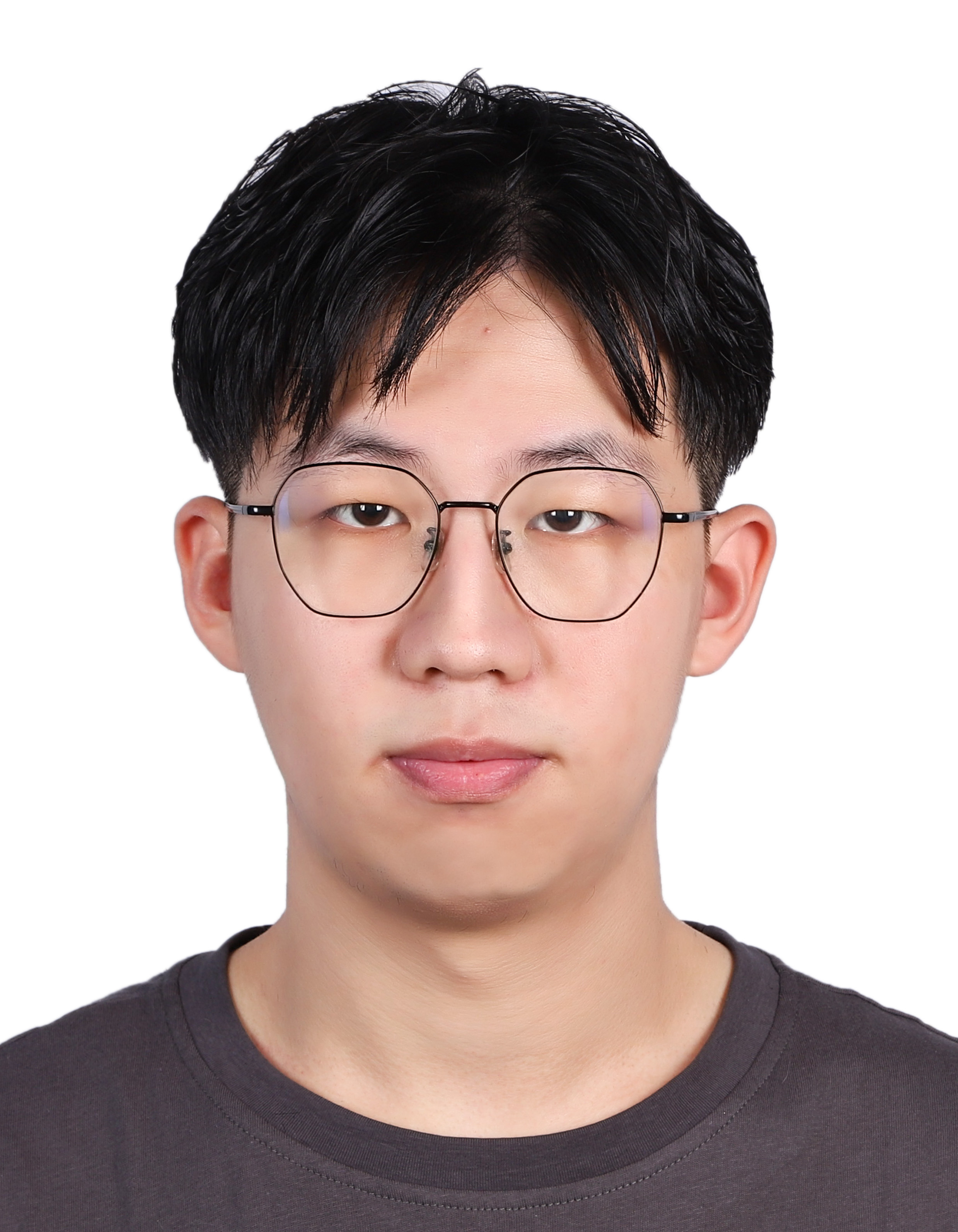}
\end{figure}

\noindent{\bf Shijie Ye}\quad is currently pursuing the B.Sc. degree in Mathematics and Statistics at the University of Toronto. He is preparing to pursue a Ph.D. degree in Artificial Intelligence, with research interests focusing on multimodal AI, large language models, deep learning, and their applications in the medical domain.

ORCID iD: 0009-0001-6681-7190.

\begin{figure}[h]%
\centering
\includegraphics[width=0.3\textwidth]{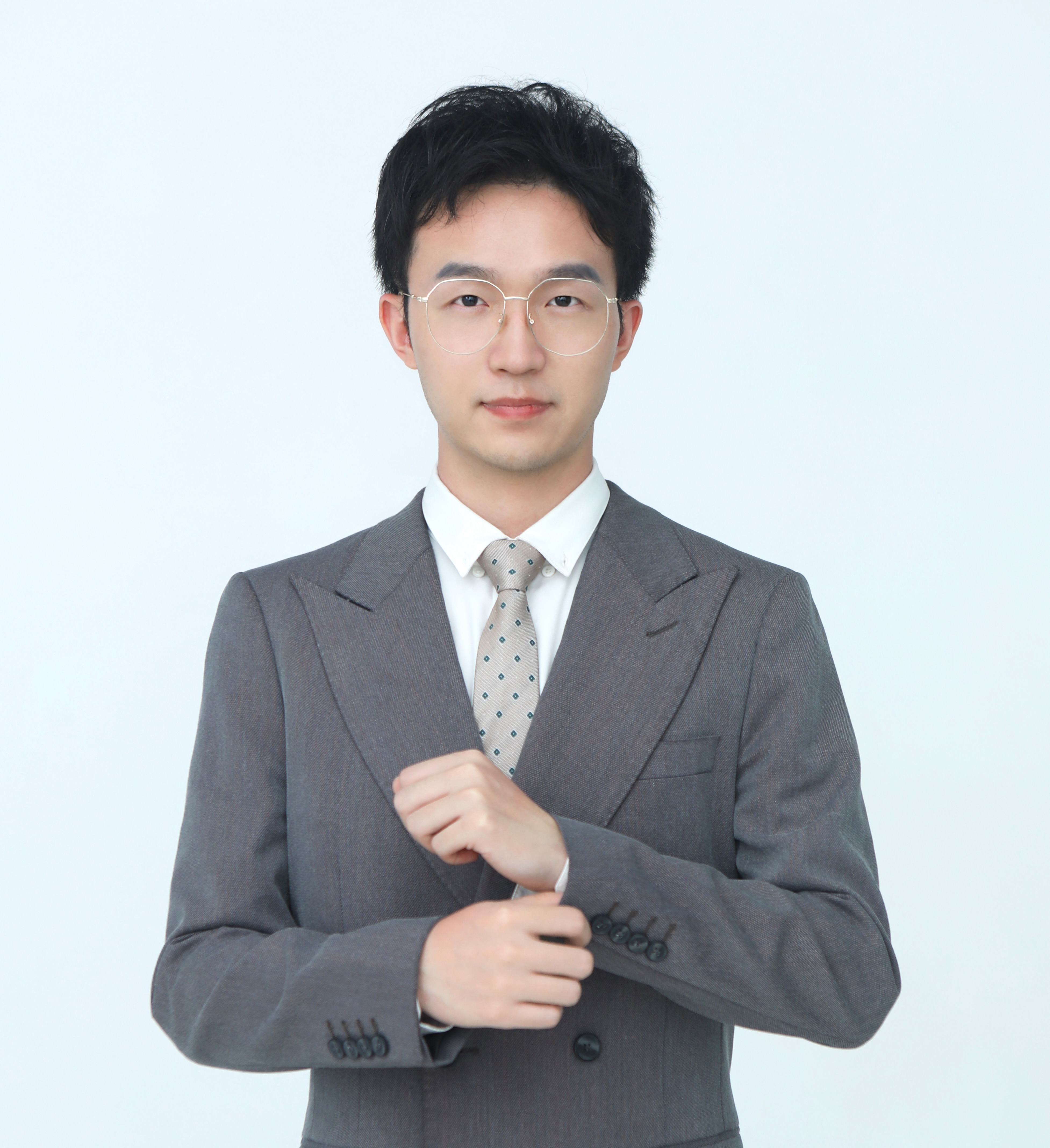}
\end{figure}

\noindent{\bf Changmeng Zheng}\quad is currently a Research Assistant Professor with the Department of Computing, the Hong Kong Polytechnic University, Hong Kong SAR. He received his Bachelor's and Master's degree from South China University of Technology, and the Ph.D. degree from the Hong Kong Polytechnic University. His research work has been published in refereed journals and conferences such as IEEE Transactions on Multimedia, IEEE Transactions on Circuits and Systems for Video Technology, Neural Networks, ACL, EMNLP, COLING, ACM MM. His research interests are in the areas of multimodal learning and social media analytics, especially the knowledge graph and large language models. 

E-mail: changmeng.zheng@polyu.edu.hk (Corresponding Author). 

ORCID iD: 0000-0002-2945-8248.

\clearpage

\begin{figure}[h]%
\centering
\includegraphics[width=0.3\textwidth]{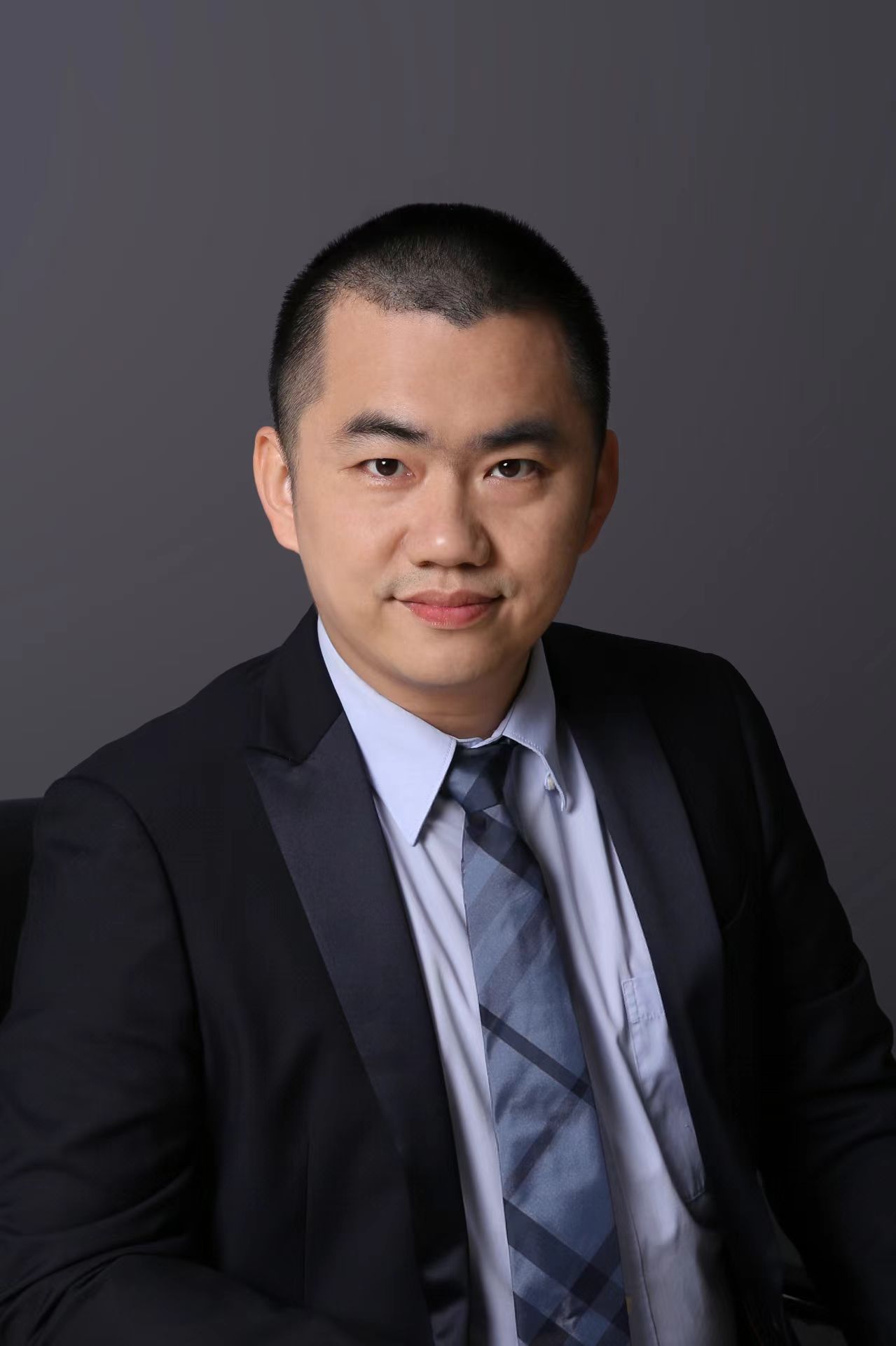}
\end{figure}

\noindent{\bf Yi Cai}, professor, serves as Dean of the School of Software Engineering at South China University of Technology. He holds concurrent appointments as Director of the Ministry of Education's Key Laboratory of Big Data and Intelligent Robotics and Director of the SCUT-Kingsoft Office Software Joint Laboratory.
As a Distinguished Member of the China Computer Federation, Professor Cai participates in several technical committees related to natural language processing and database systems. He holds leadership roles in professional organizations including the Guangdong Big Data Committee and the Standardized Knowledge Graph Alliance. Professor Cai has published more than 200 research papers in peer-reviewed journals and conference proceedings, including IEEE Transactions on Knowledge and Data Engineering and IEEE/ACM Transactions on Audio, Speech and Language Processing. His research output includes 30 granted patents. He has contributed to the development of technical standards in artificial intelligence through both national and international standardization initiatives. His research has received recognition including awards from the Geneva International Exhibition of Inventions, Guangdong Province, and the China Computer Federation. His research team has achieved competitive results in technical evaluations related to big data processing and natural language understanding.

E-mail: ycai@scut.edu.cn

\clearpage

\begin{figure}[h]%
\centering
\includegraphics[width=0.3\textwidth]{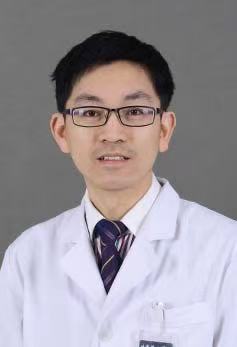}
\end{figure}

\noindent{\bf Xiaobo Yang}, M.D., Ph.D., is an Associate Chief Physician in the Department of Surgery at Peking Union Medical College Hospital (PUMCH), where he has served since 2013 after receiving his medical doctorate from Peking Union Medical College. He has been recognized as Outstanding Resident Physician and one of the “Top Ten Attending Physicians” at PUMCH, and has received multiple PUMCH Medical Achievement Awards, as well as the 3rd Prize of the Chinese Medical Science and Technology Award.

E-mail: y110403606@126.com

\begin{figure}[h]%
\centering
\includegraphics[width=0.3\textwidth]{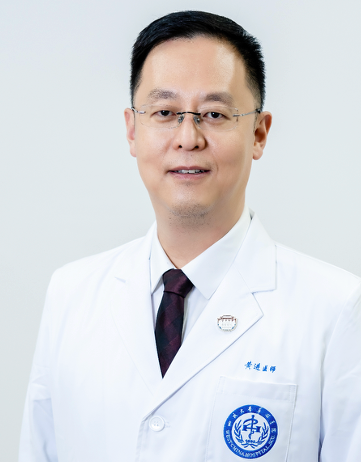}
\end{figure}

\noindent{\bf Jin Huang}, M.D. PhD, is a Professor and Doctoral Supervisor n Urology at West China Hospital, Sichuan University. Executive Vice President of West China Hospital; and Director of the Center for Medical-Engineering Innovation and Translational Research. He currently serves as Vice Chair of the Medical Engineering Branch of the Chinese Medical Association, Vice President of the China Association of Medical Equipment, among other leadership roles. He has been selected as Sichuan Provincial Academic and Technological Leader and Tianfu Qingcheng Plan Science and Technology Innovation Leading Talent. His work focuses on population-based health checkup cohorts, early-screening mechanisms for urological comorbidities, and digital-health translation. Professor Huang has led 13 national-level projects, including the National Key R\&D Program and the National Natural Science Foundation of China. He has published over 100 high-impact papers as first or corresponding author in journals such as BMJ, Nature Mental Health, and the European Journal of Epidemiology. His research was included in the BMJ “2025 China Hospital Research Impact Report” and has been cited by 55 global health policy and clinical guidelines.

E-mail: michael\_huangjin@163.com.

\begin{figure}[h]%
\centering
\includegraphics[width=0.3\textwidth]{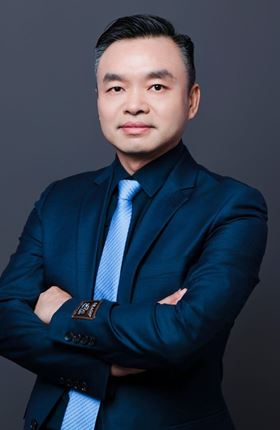}
\end{figure}

\noindent{\bf Xiaoyong Wei}\quad received the Ph.D. degree in
computer science from City University of Hong Kong, China in 2009, and has worked as a postdoctoral fellow in the University of California, Berkeley, USA from December, 2013 to December, 2015. He has been a professor and the head of Department of Computer Science, Sichuan University, China since 2010. He is an adjunct professor of Peng Cheng Laboratory, China, and a visiting professor of Department of Computing, Hong Kong Polytechnic University. He is a senior member
of IEEE, and has served as an associate editor of Interdisciplinary Sciences: Computational Life Sciences since 2020, the program Chair of ICMR 2019, ICIMCS 2012, and the technical committee member of over 20 conferences such as ICCV,
CVPR, SIGKDD, ACM MM, ICME, and ICIP. His research interests include multimedia computing, health computing, machine learning and large-scale data mining.

E-mail: x1wei@polyu.edu.hk. 

ORCID iD: 0000-0002-5706-5177.

\clearpage

\begin{figure}[h]%
\centering
\includegraphics[width=0.3\textwidth]{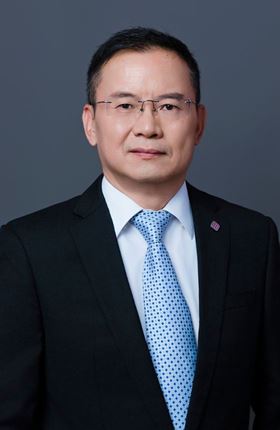}
\end{figure}

\noindent{\bf Qing Li}\quad is currently a Chair Professor (Data Science) and the Head of the Department of
Computing, the Hong Kong Polytechnic University. Formerly, he was the founding Director of the Multimedia software Engineering Research
Centre (MERC), and a Professor at City University of Hong Kong where he worked in the Department of Computer Science from 1998 to
2018. Prior to these, he has also taught at the Hong Kong University of Science and Technology
and the Australian National University (Canberra, Australia). Prof. Li served as a consultant to Microsoft Research Asia (Beijing, China),
Motorola Global Computing and Telecommunications Division (Tianjin Regional Operations Center), and the Division of Information Technology,
Commonwealth Scientific and Industrial Research
Organization (CSIRO) in Australia. He has been
an Adjunct Professor of the University of Science
and Technology of China (USTC) and the Wuhan
University, and a Guest Professor of the Hunan
University (Changsha, China) where he got his
BEng. degree from the Department of Computer
Science in 1982. He is also a Guest Professor
(Software Technology) of the Zhejiang University
(Hangzhou, China) – the leading university of the
Zhejiang province where he was born.
Prof. Li has been actively involved in the
research community by serving as an associate editor and reviewer for technical journals, and as an
organizer/co-organizer of numerous international
conferences. Some recent conferences in which
he is playing or has played major roles include
APWeb-WAIM’18, ICDM 2018, WISE2017,
ICDSC2016, DASFAA2015, U-Media2014,
ER2013, RecSys2013, NDBC2012, ICMR2012, CoopIS2011, WAIM2010, DASFAA2010, APWeb-WAIM’09, ER’08, WISE’07, ICWL’06, HSI’05,
WAIM’04, IDEAS’03,VLDB’02, PAKDD’01, IFIP
2.6 Working Conference on Database Semantics
(DS-9), IDS’00, and WISE’00. In addition, he
served as a programme committee member for
over fifty international conferences (including
VLDB, ICDE, WWW, DASFAA, ER, CIKM,
CAiSE, CoopIS, and FODO). He is currently
a Fellow of IEEE and IET/IEE, a member of
ACM-SIGMOD and IEEE Technical Committee
on Data Engineering. He is the chairperson of
the Hong Kong Web Society, and also served/is
serving as an executive committee (EXCO) member of IEEE-Hong Kong Computer Chapter and
ACM Hong Kong Chapter. In addition, he serves
as a councilor of the Database Society of Chinese
Computer Federation (CCF), a member of the
Big Data Expert Committee of CCF, and is a
Steering Committee member of DASFAA, ER,
ICWL, UMEDIA, and WISE Society.

E-mail: qing-prof.li@polyu.edu.hk.

ORCID iD: 0000-0003-3370-471X.

\end{document}